%% file: root.tex
\documentclass[letterpaper, 10 pt, conference]{ieeeconf}  

\IEEEoverridecommandlockouts                              

\usepackage{cite} 
\usepackage{multicol}
\usepackage{multirow}
\usepackage{xspace}
\usepackage{graphicx}
\usepackage{siunitx}
\usepackage{caption}
\usepackage{subcaption}
\usepackage[dvipsnames]{xcolor}
\usepackage{amssymb}
\usepackage{makecell}
\usepackage{booktabs}
\usepackage{amsmath}
\usepackage{float}
\usepackage{wrapfig}
\let\labelindent\relax
\usepackage{enumitem}
\usepackage{cuted}
\usepackage{float}
\usepackage{tabularx}
\usepackage{pifont}

\definecolor{linkblue}{RGB}{70, 130, 220}  
\usepackage[bookmarks=true, colorlinks=true, citecolor=black, linkcolor=black, urlcolor=linkblue]{hyperref}
\def\name{\mbox{MobileVISTA}\xspace}

\newcommand{\suzannah}[1]{}
\newcommand{\katliu}[1]{}
\newcommand{\stephen}[1]{}
\newcommand{\IH}[1]{}
\newcommand{\TODO}[1]{}

\title{\LARGE \bf
\name: Generative Data Augmentation \\ for Pose Generalization in Mobile Manipulation}
\author{
\textbf{Suzannah Wistreich\textsuperscript{1}, Stephen Tian\textsuperscript{1}, Isabella Huang\textsuperscript{2}, Vitor Campagnolo Guizilini\textsuperscript{2},} \\
\textbf{Sergey Zakharov\textsuperscript{2}, Katherine Liu\textsuperscript{2}, Jiajun Wu\textsuperscript{1}}\\
\textsuperscript{1}\textnormal{Stanford University} \hspace{0.05cm}
\textsuperscript{2}\textnormal{Toyota Research Institute}
}

\IEEEaftertitletext{\vspace{-2.27em}\centering\captionsetup{belowskip=0pt}
\includegraphics[width=0.91\textwidth]{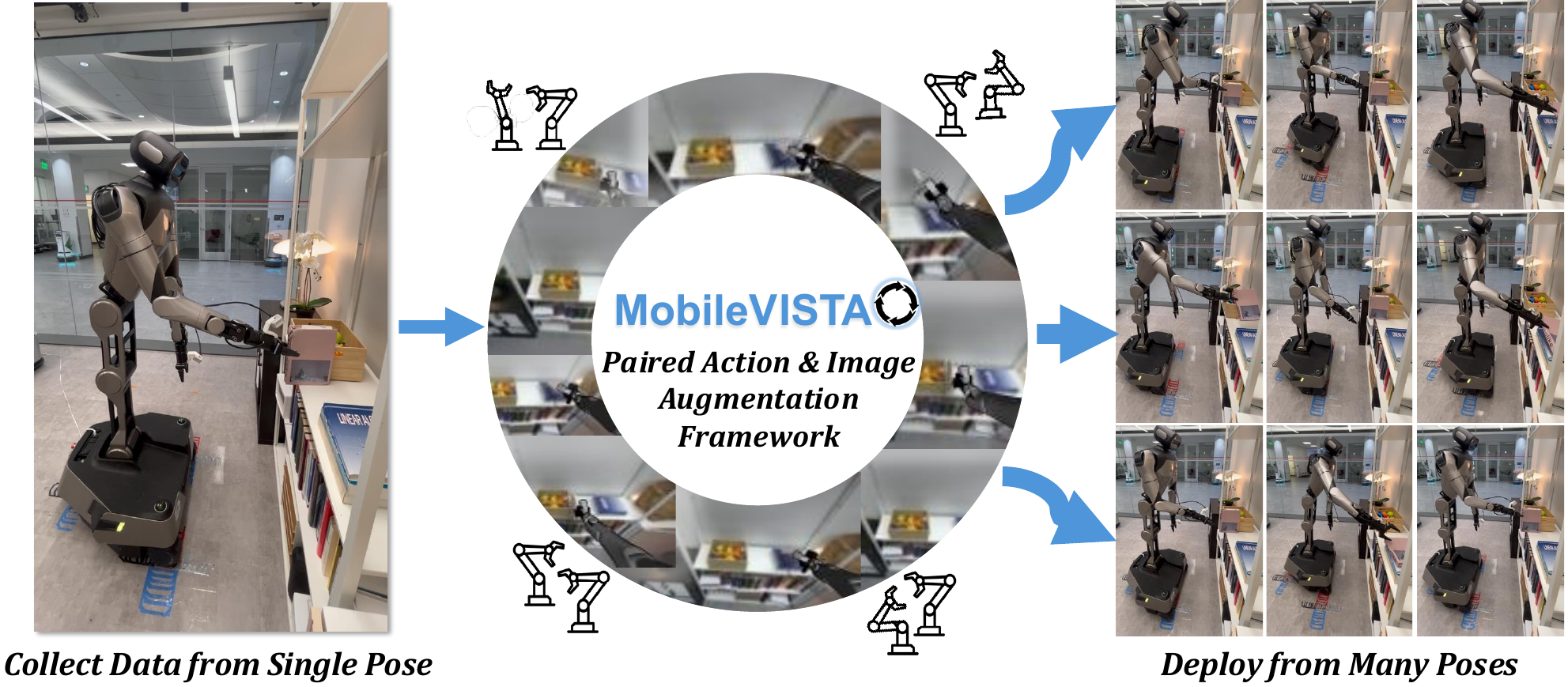}
\captionof{figure}{\textbf{MobileVISTA.} Given demonstrations collected at a single canonical robot pose (left), our framework jointly synthesizes egocentric observations and retargeted actions (center) to produce training data that allows policies to succeed from a range of unseen base poses at deployment (right).}
\vspace{0pt}}

\begin{document}

\bstctlcite{BSTcontrol}
\maketitle
\thispagestyle{empty}
\pagestyle{empty}


\begin{abstract}
\input{sec/0_abstract}
\end{abstract}

\input{sec/1_intro}
\input{sec/2_related}
\input{sec/3_framework}

\input{sec/4_experiments}
\input{sec/5_conclusion}



%


\vspace{-2mm}
\makeatletter\let\oldtb\thebibliography\def\thebibliography#1{\oldtb{#1}\setlength{\itemsep}{1.1pt}}\makeatother
\bibliographystyle{IEEEtran}
\bibliography{references}
\vspace{-10mm}
\input{sec/appendix}

\end{document}

%% file: sec/0_abstract.tex
Mobile manipulators such as humanoid robots are increasingly deployed in dynamic, unstructured environments to perform dexterous manipulation tasks.
However, end-to-end manipulation policies trained to imitate demonstration data collected from a single robot pose are brittle: even centimeter-scale deviations in robot pose at deployment can drive egocentric observations and end-effector trajectories out of the training distribution, leading to sharp drops in performance. 
We introduce \textbf{\name}, a data generation framework that transforms demonstrations captured at canonical poses into diverse, pose-perturbed training data by jointly (1) augmenting egocentric visual observations and (2) retargeting actions to compensate for base pose changes. Unlike prior methods, which assume a camera rigidly mounted off the actuated chain or non-trivial articulated robot geometry largely out of frame, \name targets compatibility with egocentric platforms (e.g., humanoids) where the camera is both influenced by \textit{and} must observe the robot's kinematic chain as it moves. We study \name in simulated tasks spanning humanoid and bimanual embodiments, and on a real Galaxea R1 Pro. We find policies trained on \name-augmented data demonstrate improved robustness to previously out-of-distribution poses encountered at test time, without additional demonstration collection or a trained generative model. Additionally, we find \name's benefit is largest on tested humanoids, where the camera rides the actuated chain and the robot fills much of the frame. Additional videos and appendix can be found on our website: \url{https://sdwistreich.github.io/mobilevista}

%% file: sec/1_intro.tex
\section{Introduction}
\vspace{-2mm}
Mobile manipulators increasingly execute complex skills in settings where navigation and manipulation are tightly sequenced, transitioning directly from locomotion to dexterous interaction from whatever configuration state navigation leaves them in. However, visuomotor policies trained to imitate a small set of task-specific demonstrations overfit to the narrow distribution of robot configurations seen during training. Large-scale pretrained policies aim to broaden this distribution~\cite{intelligence2025pi05visionlanguageactionmodelopenworld, gr00tn1_2025, generalist2025gen0}, but deploying on a new task often still relies on task-specific demonstrations, with the fine-tuned policy inheriting their restricted pose distribution.
These policies therefore fail to generalize beyond the narrow set of base poses and camera viewpoints seen during training. Even centimeter-scale base pose perturbations~\cite{yang2025mobipi}---such as those arising from reasonable navigation error---can push the robot into out-of-distribution states. 

Re-collecting demonstrations from many base poses is often impractical and scales poorly with the diversity of poses met at deployment, so a desirable property is solving tasks from whichever configuration the robot occupies \textit{without} new demonstrations. Several methods synthesize novel visual observations~\cite{tian2024vista, chen2025ropa, chen2024roviaug} to emulate camera viewpoint changes with the robot base fixed, leaving action labels unchanged. In \textit{mobile} settings, however, a base shift alters both onboard camera observations, whose viewpoints depend on the full kinematic chain from base to camera, and the actions needed to complete the task. Recent work generates the two jointly~\cite{xu2025egodemogen, xu2025r2rgen, pan20251001demos}, but assumes either a camera fixed rigidly to the base or a view containing little robot geometry beyond a gripper. Humanoids satisfy neither: the camera rides an actuated torso, so the target viewpoint is settled by the same inverse-kinematics solve that produces the retargeted action, and the arms and torso fill much of that view.



Toward addressing these limitations, we present \textbf{\name}, a data generation framework that transforms manipulation demonstrations collected from a single canonical base pose into diverse, pose-perturbed training data for mobile manipulation. Our key contributions are:
\begin{itemize}[leftmargin=1.5em, itemsep=2pt, topsep=2pt, parsep=0pt]
\item \name, a data generation framework for pose-robust mobile manipulation that produces physically consistent training data for arbitrary
base configurations entirely offline, without additional data collection.
\item An explicit geometry-based observation synthesis procedure that removes the robot from the observation, reprojects the scene with the perturbed base pose, and composites a re-rendering of the robot with retargeted joint configurations. One solve yields \textit{both} the extrinsics of the synthesized view and the robot link poses within it, separating our setting from prior work (Table I). \TODO{add ref}
\item A controlled comparison against learned novel view synthesis~\cite{ren2025gen3c}: explicit geometry-based reprojection with robot segmentation and re-compositing better preserves scene geometry and yields stronger downstream policy performance, and benefit scales with how much of the egocentric frame the robot occupies.
\end{itemize}

We evaluate \name in simulation and on a real Galaxea R1 Pro humanoid, improving over baselines that leverage 3D representations and large-scale pretrained policies under unseen base perturbations. \textbf{\name} offers a practical data generation approach for robust mobile manipulation under deployment conditions where robot pose cannot be precisely controlled.




%% file: sec/2_related.tex
\begin{figure*}[t]
\centering
\includegraphics[width=0.85\textwidth]{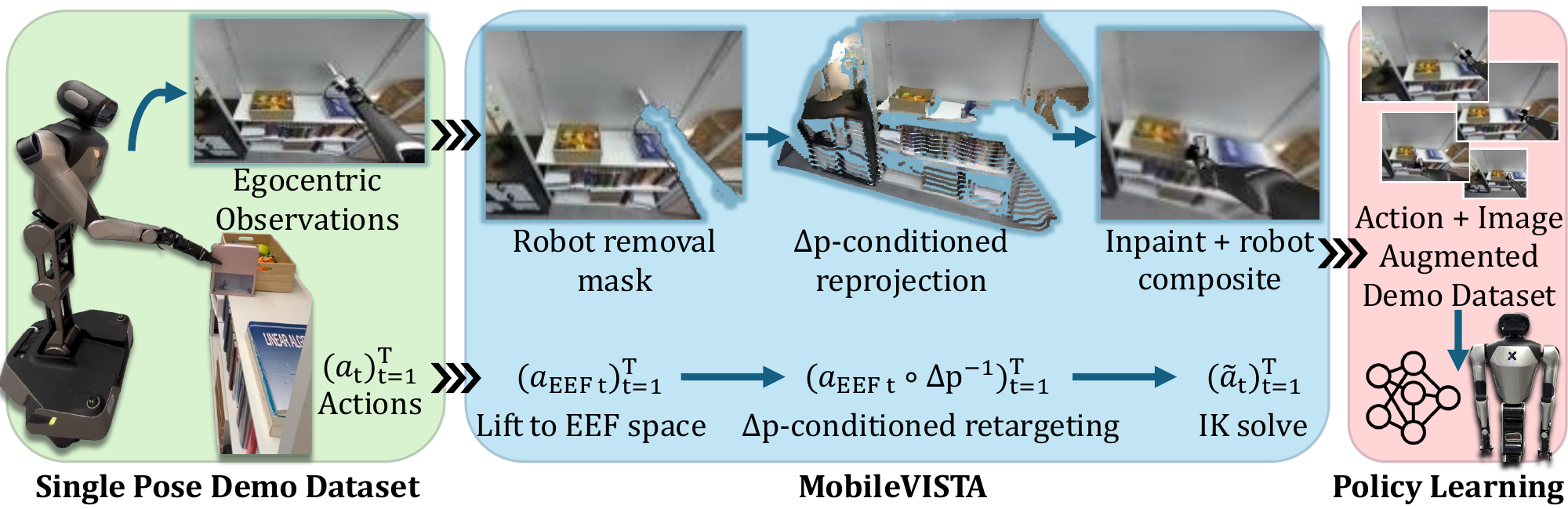}
\caption{\small \textbf{The \name framework.} Given single-pose demonstrations, \name jointly augments observations and actions under a sampled base pose perturbation $\Delta \mathbf{p} \in SE(3)$. The visual branch removes the robot, reprojects the scene, and re-renders the robot at its retargeted configuration, while the action branch retargets actions in end-effector space via $\Delta \mathbf{p}^{-1}$ to produce $(\tilde{a}_t)_{t=1}^{T}$. This augmented dataset is then used to train a pose-robust policy.}
\label{fig:framework}
\end{figure*}

\vspace{-1mm}
\section{Related Work}
\vspace{-1mm}
\textbf{Learning robust mobile manipulation.}
Learned policies for navigation and manipulation have been widely studied for mobile manipulators. End-to-end approaches train whole-body controllers that jointly produce base and arm actions~\cite{ha2024umilegs,fu2024mobile,mendonca2024continuousRL}. In contrast, modular approaches treat navigation and manipulation as sequential, decoupled components~\cite{gu2022multiskill,wu2023tidybot}. Although this ``park-and-manipulate'' paradigm is less expressive than end-to-end control, it requires less training data and is more interpretable. A key challenge in modular systems is ensuring reliable manipulation after navigation, as the post-navigation state may fall outside the manipulation policy's training distribution. ASC~\cite{yokoyama2023asc} addresses this by learning a corrective policy that adapts existing manipulation skills to such states. In contrast, Mobi-$\pi$~\cite{yang2025mobipi}, MoTo~\cite{wu2025moto}, and N2M~\cite{chai2025n2m} optimize the robot's base pose to remain within the manipulation policy’s training distribution. We instead pursue a complementary direction by designing manipulation policies that are inherently robust to a wide range of base poses.

\textbf{Synthetic data generation in robot learning.}
Synthetic data generation is a promising approach to augmenting scarce human demonstration data. Some methods generate synthetic images to improve generalization to novel object and scene configurations~\cite{mandi2022cacti, yu2023scaling, chen2023genaug, ameperosa2025rocoda,xue2025demogen}. Another application is generating cross-embodiment data, e.g. from human to robot~\cite{lepert2025phantomtrainingrobotsrobots} or robot to robot~\cite{chen2024roviaug}. Synthetic observations can also improve robustness to eye-in-hand observations~\cite{liu2025DCODA}. 

\textbf{Viewpoint-robust demonstration generation.} Several recent methods generate paired observation--action data for viewpoint robustness. RoboSplat~\cite{yang2025novel} edits a Gaussian Splatting reconstruction to vary object pose and type, appearance, lighting, and camera pose, while 1001 Demos~\cite{pan20251001demos} couples Gaussian-Splatting rendering of an eye-in-hand fisheye camera with trajectory optimization, compositing gripper geometry into rendered views. However, both assume a fixed base and a per-scene reconstruction captured in advance. VISTA~\cite{tian2024vista} and ROPA~\cite{chen2025ropa} fine-tune image diffusion models to render the scene and robot from novel third-person poses, while R2RGen~\cite{xu2025r2rgen} avoids image synthesis entirely, requiring that ``the RGB-D sensor be fixed during task execution.''
\input{table/related_comparison}

Two distinguishing properties characterize our target setting: (1) the camera rides the actuated chain, so its pose is known only once inverse kinematics resolves a configuration for the retargeted action; and (2) the robot kinematic chain is visible in frame, so synthesis must re-render articulated geometry rather than composite a rigid end-effector. 
Prior methods satisfy at most one (Table~\ref{tab:related_comparison}). Closest to our setting is EgoDemoGen~\cite{xu2025egodemogen}, generating paired egocentric observations and actions under planar base perturbations, but whose ``egocentric camera is rigidly mounted on the robot head, which is fixed to the base.'' \name instead targets cameras on the actuated chain, and synthesizes via explicit geometry rather than a per-domain fine-tuned video generator.

We do not run the methods of Table~\ref{tab:related_comparison} as baselines. Five methods assume a fixed base and therefore define no action at a perturbed pose. Of the two that consider a mobile base, R2RGen synthesizes no images, and EgoDemoGen, which synthesizes with a learned video generator, has no public implementation as of submission. We instead compare against learned NVS directly, using GEN3C (Section~\ref{sec:gen3c}), and isolate the contribution of robot re-rendering through a \textit{Fixed Robot Geom.} ablation (Section~\ref{sec:ablation}).

%% file: table/related_comparison.tex
\begin{table}[t]
\centering
\small
\setlength{\tabcolsep}{1.5pt}
\begin{tabularx}{\linewidth}{@{}>{\hsize=1.10\hsize\raggedright\arraybackslash}X>{\hsize=0.90\hsize\raggedright\arraybackslash}Xccc@{}}
\toprule
\textbf{Method} & \textbf{Embodiment} & \makecell{Mobile\\base} & \makecell{Camera on\\act.\ chain} & \makecell{Robot vis.\\past gripper} \\
\midrule
VISTA~\cite{tian2024vista} & Single-arm & \ding{55} & \ding{55} & \ding{51} \\
RoVi-Aug~\cite{chen2024roviaug} & Single-arm & \ding{55} & \ding{55} & \ding{51} \\
ROPA~\cite{chen2025ropa} & Biman. & \ding{55} & \ding{55} & \ding{51} \\
RoboSplat~\cite{yang2025novel} & Single-arm & \ding{55} & \ding{55} & \ding{51} \\
EgoDemoGen~\cite{xu2025egodemogen} & Biman. & \ding{51} & \ding{55} & \ding{51} \\
1001 Demos~\cite{pan20251001demos} & Single-arm & \ding{55} & \ding{51} & \ding{55} \\
R2RGen~\cite{xu2025r2rgen} & Single, biman. & \ding{51} & \ding{55} & --- \\
\midrule
\textbf{\name} (ours) & \textbf{Biman., humanoid} & \ding{51} & \ding{51} & \ding{51} \\
\bottomrule
\end{tabularx}
\caption{\small \textbf{Settings targeted by prior work.} Embodiment lists every platform class each method is evaluated on, in simulation and on hardware. Only two prior methods consider a mobile base, neither carries the camera on the actuated chain, and none is evaluated on a humanoid. R2RGen synthesizes no images (---).}
\label{tab:related_comparison}
\vspace{-3mm}
\end{table}

%% file: sec/3_framework.tex
\vspace{-1mm}
\section{The \name Framework}
\label{sec:framework}
\vspace{-2mm}
Given a dataset of expert demonstrations collected from a single canonical robot base pose, \name transforms each demonstration into pose-perturbed variants by modifying both egocentric visual observations and their corresponding actions. Training on augmented data exposes policies to diverse robot configurations, enabling them to handle previously out-of-distribution poses at test time. Figure~\ref{fig:framework} illustrates the method.

Each demonstration consists of time-aligned egocentric and wrist camera RGB-D images, proprioceptive robot states, and joint-space actions including gripper and finger actuation. Joint-space configurations can be mapped to end-effector poses via forward kinematics. Concretely, given a source trajectory $\tau = {(o_t, a_t)}_{t=1}^{T}$ collected at canonical base pose $P \in SE(3)$, \name synthesizes a pose-perturbed trajectory $\tilde{\tau} = {(\tilde{o}_t, \tilde{a}_t)}_{t=1}^{T}$ under randomly sampled pose transformations $P \circ  \Delta \mathbf{p}$, where $\tilde{o}_t$ are pose-consistent rendered observations and $\tilde{a}_t$ are retargeted actions aimed to preserve the original end-effector trajectory known to successfully complete a target task. For each synthesized episode, we sample a target base translation offset $\Delta \mathbf{b} \in \mathbb{R}^2$ 
uniformly within a disk of radius $r$ centered at the canonical pose and a yaw rotation perturbation $\Delta \theta \sim \mathcal{U}(-5^\circ, 5^\circ)$. Together, these define the pose perturbation $\Delta \mathbf{p} \in SE(3)$, the transformation  induced by $(\Delta \mathbf{b}, \Delta \theta)$. 
Given these, \name jointly augments (i) visual observations by synthesizing pose-shifted egocentric views of the scene and (ii) action labels by retargeting the original motion to compensate for the base perturbations. 
We leave wrist views unmodified: their localized viewpoints and the sparse depth of  close-range interaction make geometry-based reprojection unreliable, and as retargeting aims to preserve original end-effector trajectories, wrist-camera observations can stay largely consistent under the base perturbation.



\subsection{Generating Visual Observations} 
\label{sec:vis-obs}
\vspace{-2mm}
To generate pose-perturbed visual observations consistent with a target base offset, we assume access to a simulation model of the robot geometry. The model serves three purposes: (1) segmenting the robot out of the original egocentric image, (2) re-rendering the robot under retargeted actions to composite with the synthesized scene, and (3) recovering the egocentric camera trajectory those actions induce along the kinematic chain from base to camera.

We adopt a geometry-based segmentation-and-reprojection pipeline to synthesize new egocentric observations. We first render the robot model using the canonical action trajectory and camera parameters to obtain a precise 2D robot geometry mask. Applying this mask to the egocentric depth image removes the robot, yielding a robot-free depth map of the scene. We then de-project the robot-free RGB-D image into a colored point cloud using a pinhole camera model. To account for camera motion induced by retargeting (e.g., torso or waist adjustments), we lift the point cloud into the world frame using the original per-timestep camera simulator extrinsics obtained while rendering the original actions. We apply the target base perturbation in world coordinates, and then re-project the transformed point cloud from the \textit{new} egocentric camera, whose per-timestep extrinsics are obtained by executing the retargeted actions with the robot model. Note that egocentric camera poses are not chosen independently of actions\IH{I don't know if this is explained clearly enough. It doesn't seem like the second part of this sentence says anything about independence of camera pose from the actions.}: while a perturbed base pose may allow multiple valid camera poses when intermediate joints are unconstrained, camera extrinsics are inherited from the joint configuration resolved by the retargeted actions (Section~\ref{sec:retargeting}) via IK. \IH{One question that pops up while reading this-- what happens if there are multiple valid egocentric camera poses for the same perturbed base pose? This is often the case when the robot has a torso or lift joint.} \suzannah{updated to mention that camera inherits pose from resolved actions from IK} \IH{I guess my original question was moreso like, 'what if there are multiple IK solutions for the camera pose?' i.e., does it matter that we're only choosing one valid camera pose?} 

\begin{figure*}[t]
    \centering
\includegraphics[width=0.98\textwidth]{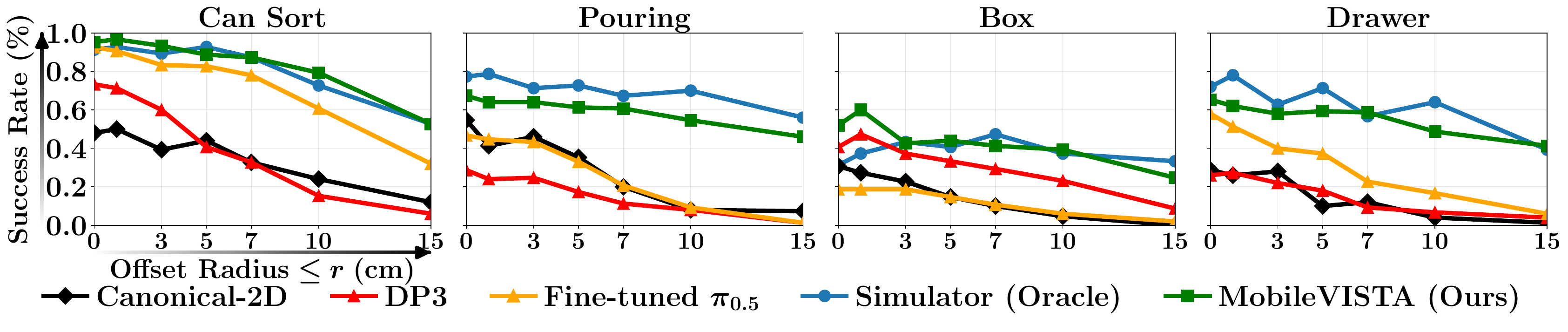}
    \caption{\small \textbf{Randomized pose perturbation results.} Success rate as a function of evaluation perturbation radius $r_{eval}$ (with yaw $\Delta \theta \sim \mathcal{U}(-5^\circ, 5^\circ)$). \name maintains higher success rates than all baselines as perturbation magnitude grows.}
    \label{fig:randomized}
\end{figure*}
This produces a pose-shifted reprojection of the static environment, but introduces missing regions due to both the removal of the original robot \textit{and} newly exposed areas from the scene reprojection. As in prior work~\cite{lepert2025phantomtrainingrobotsrobots}, we fill these holes, along with any regions missing due to incomplete depth, using the image inpainting model E\textsuperscript{2}FGVI~\cite{liCvpr22vInpainting}. We inpaint rather than nearest-pixel filling, since disocclusions expose unobserved content near boundaries rather than repeatable texture. Finally, after computing retargeted actions, we render the robot model executing the retargeted actions to obtain accurate link poses from the egocentric camera, and composite it with the synthesized background image. The same pipeline runs on real demonstrations, with the robot mask obtained from a rendered URDF depth buffer and intrinsics taken from the ZED2's calibration (Appendix~\ref{sec:appendix_visual_synthesis}). It is fully automatic, with no manual filtering at any stage.

\subsection{Retargeting Actions}
\label{sec:retargeting}
\vspace{-2mm}
We retarget actions to compensate for robot base pose perturbations in end-effector space. When source actions are joint-space, they are first converted into robot base frame end-effector goals via forward kinematics. We then apply the inverse of pose perturbation to each per-timestep end-effector goal, approximately preserving the \textit{original} end-effector trajectory that is known to successfully complete the task. Where required, retargeted end-effector goals are converted back into joint-space actions via whole-body inverse kinematics. End-effector actuation (e.g., gripper or finger commands) is left unchanged.

Simulated demonstrations are recorded as end-effector poses, making retargeting closed-form, while JoyLo teleoperation records joint-space commands, which require an IK solve. The IK solve enforces smoothness, velocity bounds, joint limits, and self-collision avoidance (Appendix~\ref{sec:appendix_action_retargeting}). Past the initial transient, the retargeted trajectory tracks the original world-frame end-effector path, which is collision-free by construction. Within the sampled $15$cm range all trajectories converged, so we apply no filtering. Under significant perturbations, IK failure would largely coincide with the task being unreachable from that base pose---a case for repositioning rather than augmentation. As in behavior cloning generally, we assume task-relevant contact and object dynamics are approximately unchanged under the sampled perturbations.

%% file: sec/4_experiments.tex
\section{Experiments}
\vspace{-2mm}
In this section, we perform empirical analyses to answer the following questions:

\begin{list}{}{\leftmargin=1.5em \itemsep=2pt \topsep=0pt \parsep=0pt}
\item[] \textbf{Q1:} Does \name improve robustness to base poses outside the training distribution at test time, across humanoid and non-humanoid embodiments?

\item[] \textbf{Q2:} Are both coupled visual–action augmentation and explicit robot re-rendering at retargeted configurations necessary for pose generalization? 

\item[] \textbf{Q3:} Does \name outperform learned novel view synthesis for policy pose robustness?

\item[] \textbf{Q4:} Is \name effective when deployed on real robotic hardware?

\end{list}

We compare \name with the following baselines and prior methods:
\vspace{-1mm}
\begin{itemize}[leftmargin=1.5em]
    \item \textbf{Canonical-2D.} Trains diffusion policies~\cite{chi2023diffusionpolicy} on 2D image observations from the original single canonical base pose only, without pose or viewpoint augmentation.
    
    \item \textbf{Simulator (Oracle).} Relocates the robot base to the sampled perturbed pose in simulation, retargets actions as in Section~\ref{sec:retargeting}, and records the egocentric observations from executing these actions. This yields ground-truth observations exactly consistent with the retargeted actions, upper-bounding \name by testing perfect synthesis. This has no real-world counterpart, as re-recording demonstrations at every perturbed pose is not tractable at scale. 
    \item \textbf{DP3~\cite{Ze2024DP3}}. Trains policies on explicit 3D scene representations (e.g., point clouds) rather than RGB images, isolating the effect of geometry-aware inputs on pose robustness. Like Canonical-2D, DP3 trains only on canonical demonstration data, differing solely in input representation.
    \item \textbf{Fine-tuned $\boldsymbol{\pi}_{0.5}$~\cite{intelligence2025pi05visionlanguageactionmodelopenworld}}. Fine-tunes pretrained $\pi_{0.5}$ policy on each task's canonical demonstration data, without pose augmentation or action retargeting, testing whether large-scale pretraining alone suffices for pose generalization.
\end{itemize}

Note that \name is a data generation framework, not a policy architecture: \name, Canonical-2D, and every ablation uses \textit{identical} Diffusion Policy \cite{chi2023diffusionpolicy} architecture, hyperparameters (Appendix Table~\ref{tab:dp_hyperparams}) and pipeline, differing only in training data, with no architectural change needed to consume augmented data. DP3 and fine-tuned $\pi_{0.5}$ instead test whether an alternative input representation alone confers pose robustness. Simulator (Oracle) matches \name in architecture, data volume, sampled offsets, and training steps. Throughout, we use $r_{\text{train}} = r_{\text{eval}} = 15$cm and $\Delta \theta \sim \mathcal{U}(-5^\circ,5^\circ)$, chosen to span realistic post-navigation base error~\cite{yang2025mobipi}. Source-to-augmented ratios are 1:1 in simulation and 1:10 in real, and each simulated result is a mean over three seeds at 50 rollouts.

\begin{figure*}[t]
    \centering
    \includegraphics[width=0.98\textwidth]{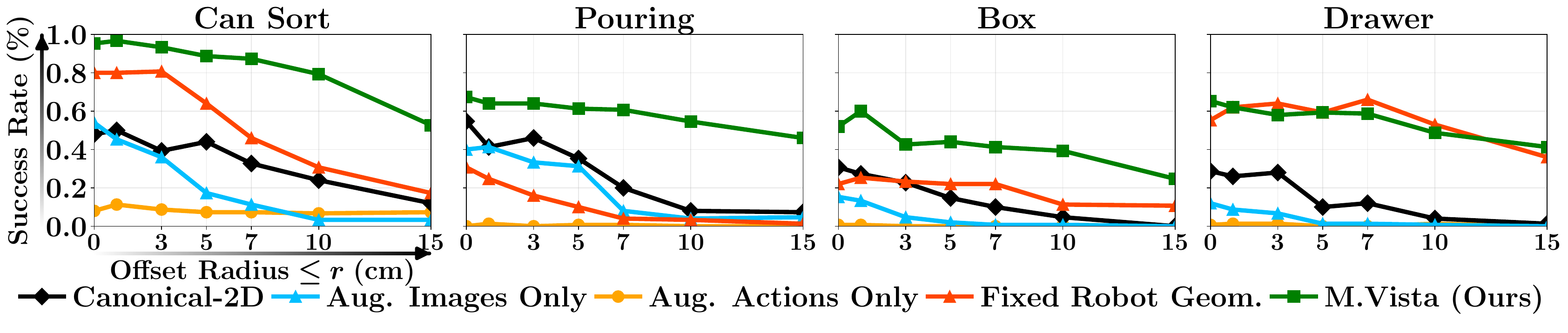}
    \caption{\small \textbf{Ablation results across tasks.} Success rate vs. evaluation perturbation radius $r_{eval}$. We compare \name against single-modality augmentation---\textit{Aug. Images Only, Aug. Actions Only}--- and an augmentation that omits robot re-rendering at retargeted configurations, \textit{Fixed Robot Geom.} \name outperforms nearly all ablated variants, indicating that both coupled visual–action augmentation and explicit robot re-rendering contribute to pose robustness.}
    \label{fig:ablation}
\end{figure*}

\subsection{Simulated Experimental Setup}
\label{sec:sim}
\vspace{-2mm}
We use the robomimic framework with the MuJoCo simulator \cite{mujoco} on tasks from DexMimicGen~\cite{jiang2024dexmimicen}. Can Sort trains on an even split of expert demonstrations across can color; the remaining tasks train on the first 200 demonstrations and evaluate on the DexMimicGen environment variants. Can Sort and Pouring tasks are performed on a Fourier GR-1 humanoid platform, while Box and Drawer tasks are evaluated on a bimanual Franka setup. Note that our simulated experiments serve to isolate synthesis quality against an oracle requiring simulator access, which has no real hardware equivalent. The benefit of augmenting over manual collection is tested on real hardware (Section~\ref{sec:real}).

Per trial, the robot base is displaced by an offset sampled uniformly within a disk of radius $r_{eval}$ about the canonical pose, with yaw $\Delta \theta \sim \mathcal{U}(-5^\circ, 5^\circ)$ held fixed across all $r_{eval}$, including $r_{eval} = 0$. We sweep performance across radius sampling range $r_{eval} \in [0\mathrm{cm}, 15\mathrm{cm}]$, evaluating performance on progressively larger out-of-distribution base perturbations.

\textbf{Q1: Robustness to Out-of-Distribution Base Poses.} Results are shown in Figure~\ref{fig:randomized}. MobileVISTA consistently outperforms all baselines as the randomization range grows. In contrast, Canonical-2D 
training degrades rapidly as $r$ increases, averaging under 45\% success at $r_{eval} = 3\mathrm{cm}$ across all tasks. DP3 and fine-tuned $\pi_{0.5}$ match or modestly improve on Canonical-2D in most cases, but demonstrate similarly limited generalization as perturbation magnitude increases. Notably, fine-tuned $\pi_{0.5}$ achieves robustness close to \name and Simulator (Oracle) in the Can Sort task for offsets less than $r_{eval}=5\mathrm{cm}$. Overall, Canonical 2D training, explicit 3D representations, and large pretrained initialization are each insufficient on their own for robust pose generalization. These trends hold on both the humanoid GR-1 (Can Sort, Pouring) and the non-humanoid bimanual Franka (Box, Drawer). Full tabled results are in Appendix Section \ref{sec:appendix_detailed_results}. 

\subsection{Ablation Studies}
\label{sec:ablation}
\vspace{-2mm}

\textbf{Q2: Contribution of \name's Components.} We ablate key components of MobileVISTA to understand their contributions. We evaluate whether: (1) \textit{coupled} visual–action augmentation is necessary for pose generalization, and (2) explicitly re-rendering the robot at retargeted joint configurations is required to synthesize in-distribution egocentric observations.

\subsubsection{Coupled Visual--Action Augmentation}
\label{sec:single-mod}

Some augmentation frameworks~\cite{tian2024vista, chen2025ropa, chen2024roviaug} 
perturb visual observations in isolation, assuming the action modality can be treated as fixed. To test whether this suffices for mobile manipulation, we pair the MobileVISTA-augmented data from Section~\ref{sec:sim} with its canonical source data, preserving demonstration order: \textit{Aug. Actions Only} pairs canonical observations with retargeted actions, and \textit{Aug. Images Only} pairs augmented observations with the original canonical actions, so exactly one modality is augmented relative to its source demonstration. 
Results are shown in Figure~\ref{fig:ablation}. We find that augmenting either modality in isolation consistently degrades performance across tasks, showing that jointly augmenting visual observations and actions is critical.

\subsubsection{Robot-Consistent Rendering After Retargeting}

MobileVISTA explicitly re-renders the robot at retargeted joint configurations, compositing this overlay onto the reprojected environment to ensure visual observations remain consistent with executed actions. While Section~\ref{sec:single-mod} examines whether both modalities must be augmented, here we examine whether the visual branch must account for the robot's \textit{own} motion under the base shift, rather than treating its geometry as static. \IH{Does Aug. Actions in 4.2.1 use robot-consistent rendering? Not super clear to me.} We ablate our pipeline's robot-consistent rendering by omitting 
the robot segmentation and rendering overlay of Section~\ref{sec:vis-obs}: the \textit{entire} scene is re-projected from depth to the target view without segmenting the robot out, treating it as static in the environment rather than coupled with the egocentric camera perturbation. 
 \begin{figure*}[t]
    \centering
    \includegraphics[width=0.98\linewidth]{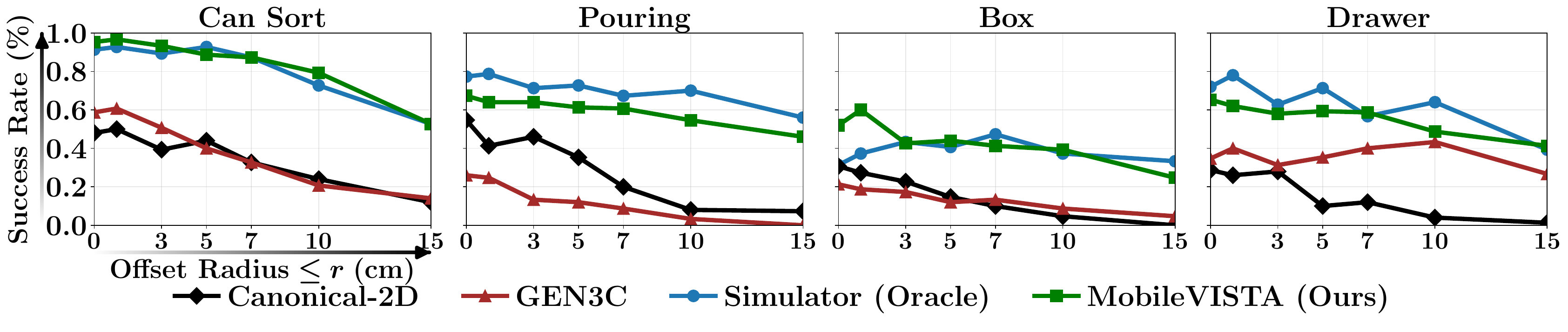}
    \caption{\small \textbf{Policy performance under \name vs. GEN3C-synthesized training data.} Success rate vs. evaluation perturbation radius $r_{eval}$, alongside the Simulator (Oracle) upper bound. Policies trained on \name-synthesized data outperform those trained on GEN3C-synthesized data across tasks and perturbation magnitudes.}
    \label{fig:gen3c_comparison}
\end{figure*} Results are shown in Figure~\ref{fig:ablation}. Performance drops substantially under this setting in most tasks, suggesting that explicitly rendering accurate robot geometry after retargeting is critical for maintaining in-distribution egocentric observations. 

\textbf{When does robot-consistent rendering matter?} Averaged over radii, \name's gap over \textit{Fixed Robot Geom.} is largest on Pouring ($+0.47$), smaller on Can Sort and Box ($+0.28$ and $+0.24$), and absent on Drawer ($\approx 0.00$). We postulate robot-consistent rendering matters most under two task properties: how much of the egocentric frame the robot's geometry occupies, and whether the camera rides the kinematic chain. Pouring, on the humanoid GR-1 with its actuated waist, has both and shows the largest gap, while the effect is weakest on the Franka tasks, where the rig-mounted camera sees less of the robot. This remains a hypothesis, as four tasks cannot separate the two properties and do not account for the Box--Drawer difference.

\vspace{-1mm}
\subsection{Quality Analysis of Synthesized Observations}
\label{sec:gen3c}
\setlength{\intextsep}{0pt}
\vspace{-2mm}


\textbf{Q3: Comparison of \name with learned novel view synthesis.} \name generates pose-consistent observations from explicit geometry; learned novel view synthesis (NVS) is an alternative. 
We compare \name-generated egocentric views against those produced by a representative NVS model via controlled PSNR evaluation. We render oracle views from the simulator at perturbed base poses and treat these as ground truth, then measure PSNR between these oracle images and views synthesized by (i) MobileVISTA, and (ii) GEN3C~\cite{ren2025gen3c}, a leading learned NVS model for 3D-consistent camera-controlled video generation, under identical camera perturbations. 


We train policies on the GEN3C views with the same action retargeting (Section~\ref{sec:retargeting}), and compare performance against \name and oracle policies of Section~\ref{sec:sim}. 

\begin{figure}
    \centering
    \includegraphics[width=0.76\linewidth]{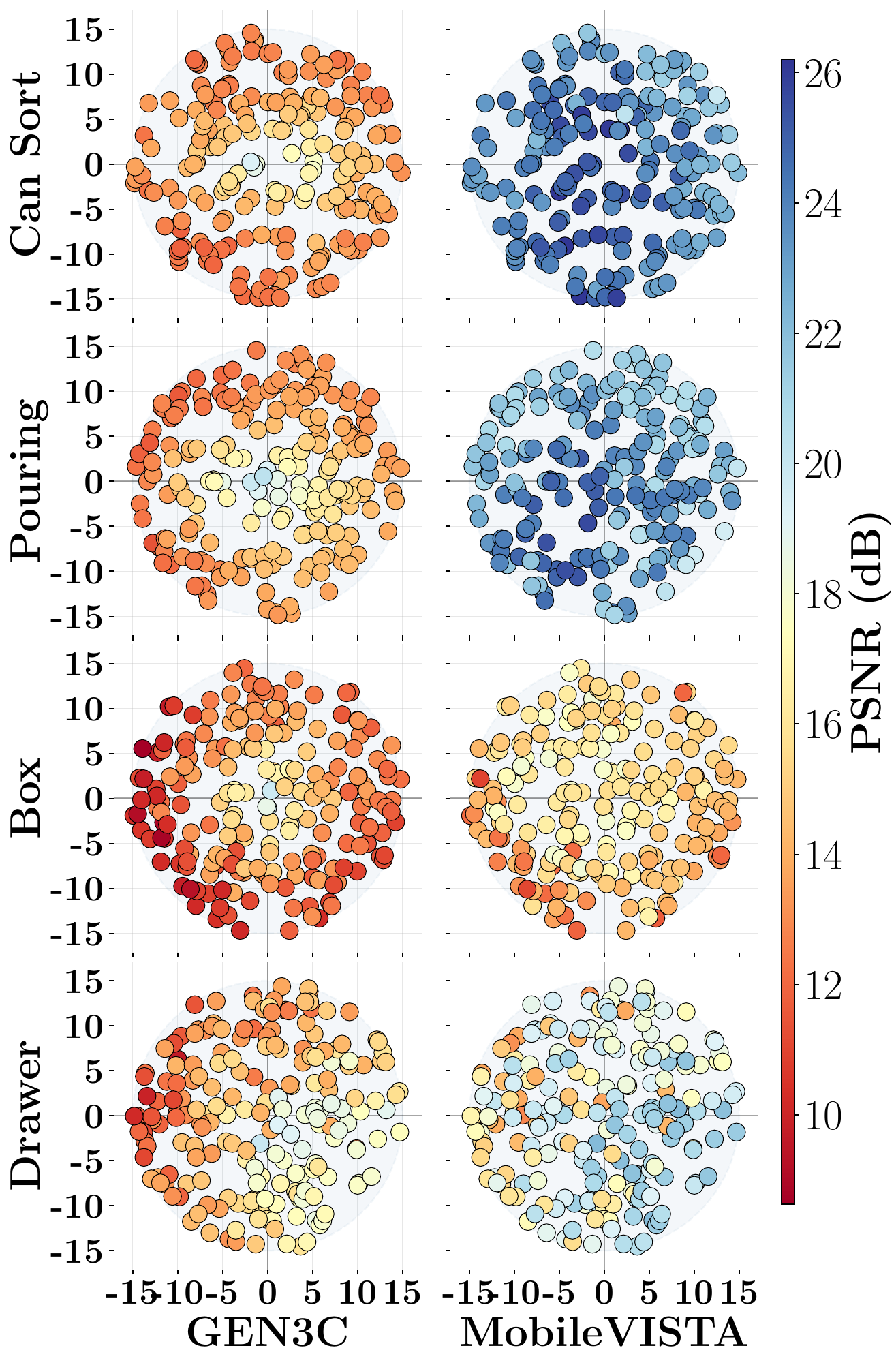}
    \caption{\small \textbf{Per-trajectory PSNR for synthesized images.} Each point shows the PSNR ($\uparrow$, dB) of a synthesized trajectory at its sampled base offset from canonical (cm, overhead view). \name yields higher PSNR overall, with the largest separation on humanoid tasks. Tasks where the methods are closer in PSNR also show smaller gaps in downstream policy performance.}
    \label{fig:psnr_heatmap}
\end{figure}

As shown in Figure~\ref{fig:psnr_heatmap}, MobileVISTA consistently achieves higher PSNR across perturbation magnitudes, indicating more consistent preservation of scene geometry and object structure. In contrast, GEN3C synthesis degrades noticeably under larger viewpoint shifts, often producing blurred object boundaries or distorted geometry. The gap is not uniform across embodiments: at offsets of $10$--$15$cm, \name exceeds GEN3C by $10.2$dB and $8.9$dB on the two humanoid GR-1 tasks, against $2.9$dB and $3.9$dB on the two Franka tasks (Appendix Table~\ref{tab:nvs_psnr}). Results from policy learning are presented in Figure~\ref{fig:gen3c_comparison}. We find that policies trained on \name-synthesized data outperform those trained on GEN3C across tasks and perturbation magnitudes, consistent with \name's higher reconstruction fidelity. 

\begin{figure*}[t]
    \centering
    \begin{subfigure}{\textwidth}
        \centering
        \includegraphics[width=0.82\linewidth]{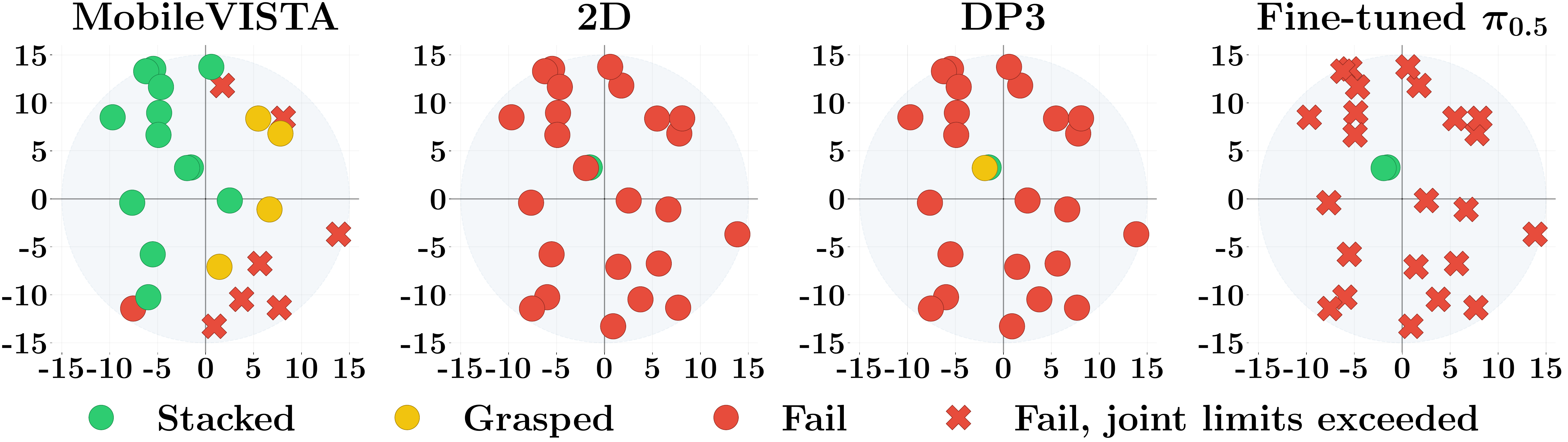}
        \caption{\textbf{Book Stacking}}
        \label{fig:real}
    \end{subfigure}
    \\[15pt]
    \begin{subfigure}{\textwidth}
        \centering
        \includegraphics[width=0.82\linewidth]{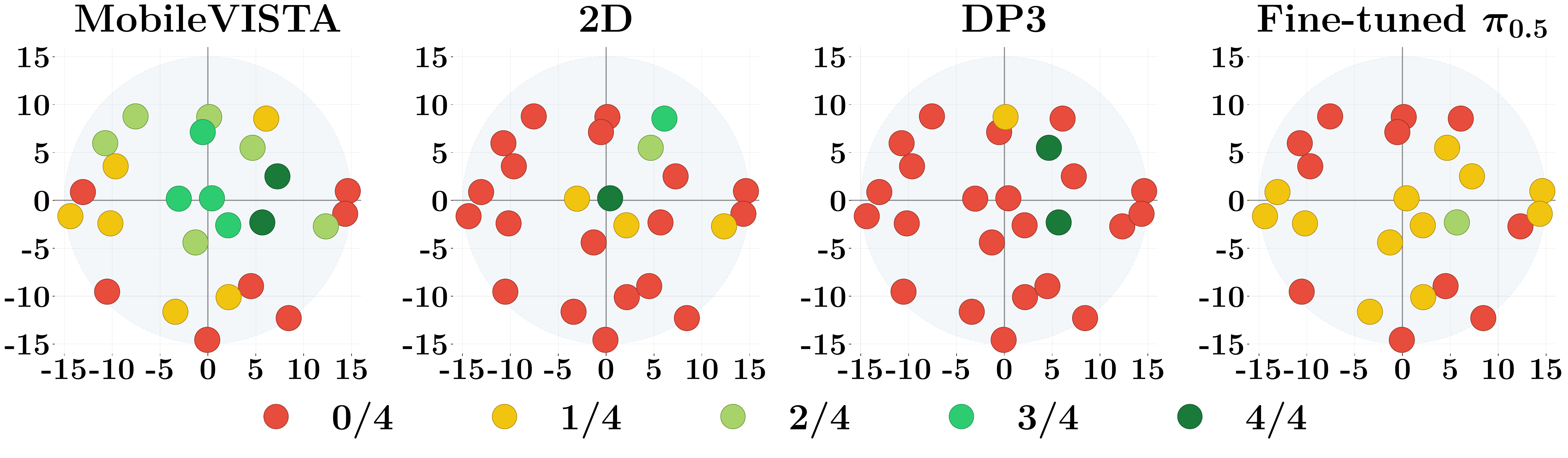}
        \caption{\textbf{Laundry Sort}}
        \label{fig:laundry_real_plot}
    \end{subfigure}
    \vspace{0.mm}
    \caption{\small \textbf{Sampled offsets at $\boldsymbol{r_{eval}=15}$\textbf{cm} in real-world evaluation.} Outcomes for each policy across sampled base offsets (cm, overhead view), for book stacking (a) and laundry sort (b). \name succeeds across a wide range of offsets, while baselines fail to generalize beyond the canonical pose.}
    \label{fig:real_combined}
\end{figure*}

Additionally, \name generates augmented data 4-7$\times$ faster than GEN3C across tasks, with full per-task averages tabled in Appendix Table~\ref{tab:timing}. Full tabled PSNR values, as well as qualitative comparisons to further highlight these effects, can be found in Appendix Section~\ref{sec:appendix_psnr}.

Beyond reconstruction quality, NVS-based approaches treat the robot as part of the static scene, limiting how well it reflects base-conditioned changes in robot configuration. Prior work ~\cite{tian2024vista} showed promising NVS results for externally mounted cameras observing static robots; we postulate the static-robot assumption is especially problematic in egocentric settings, where camera motion and robot articulation are tightly coupled. \name instead re-renders the robot under retargeted actions and composites it with a geometry-consistent background. 

\subsection{Real-World Evaluation}
\label{sec:real}
\vspace{-2mm}
\textbf{Q4: Deployment on real hardware.} We investigate the performance of \name on real-robot data using a Galaxea R1 Pro humanoid. We evaluate on two representative real-world manipulation tasks:
\begin{itemize}[leftmargin=*,noitemsep,topsep=1pt]
    \item \textbf{Laundry Sort}: the robot must sort four laundry items into two bins based on category. This task is challenging due to its long horizon, the deformability of the clothing items, and the bimanual coordination required to sort multiple objects in sequence.
    \item \textbf{Book Stacking}: the robot must grasp a vertically placed book and reorient it horizontally onto an existing stack inside a bookshelf. This is challenging due to the precise lateral alignment required for a successful grasp.
\end{itemize}
We collect $50$ expert demonstrations per task from a non-perturbed robot pose using the JoyLo framework introduced in BEHAVIOR Robot Suite~\cite{jiang2025brs}, and train a \name policy and baselines using the same framework as described in Section~\ref{sec:framework}.

\textbf{Evaluation procedure.} For each trial, we sample a base offset uniformly within $r_{eval}=15\mathrm{cm}$ of the canonical pose and manually drive the robot to that offset. Yaw is not explicitly controlled, as policies are trained over the full $\pm5^\circ$ yaw range used during augmentation. From each sampled offset, we roll out all four policies (\name, Canonical-2D, DP3, Fine-Tuned $\pi_{0.5}$) without repositioning the robot. The sampled offset is held fixed across all four policies with no repositioning between them, so docking stochasticity cannot advantage any
method, and policy evaluation order at each offset is randomized and unknown to the experimenter scoring the rollout. We score each Book Stacking rollout in two stages: \textit{Grasped}, where the robot successfully grasps the book, and 
\textit{Stacked}, where the book is successfully placed onto the existing stack.

\textbf{Results.} Full success rates are reported in Table~\ref{tab:eval} and the spatial distribution of outcomes at $r_{eval}=15\mathrm{cm}$ in Figure~\ref{fig:real_combined}. On Book Stacking, the \name-trained policy grasps in 17/25 randomized trials and stacks in 13/25, while no baseline exceeds 2/25 at either stage. 

\input{table/real}

\name generalizes particularly well to leftward and forward offsets; its remaining failures are predominantly joint-limit violations when reaching for the book, which we postulate arise when the retargeted trajectory requires the arm to cross the torso. In contrast, Canonical-2D, DP3, and Fine-Tuned $\pi_{0.5}$ execute the original demonstration trajectory without compensating for the new base pose: forward offsets cause collisions with the shelf, backward offsets leave the gripper short of the book. Fine-Tuned $\pi_{0.5}$ also consistently outputs actions outside the robot's joint limits, though its qualitative failure modes resemble Canonical-2D and DP3.

On Laundry Sort, \name sorts at least one item in 18/25 trials and at least two in 12/25; the strongest baselines reach 14/25 (fine-tuned $\pi_{0.5}$) and 3/25 (Canonical-2D). Failure modes separate by method and direction. Qualitatively, Canonical-2D and DP3 policies do not compensate for backward shifts, stopping short when attempting to grasp items. Canonical-2D additionally executes the bimanual handoff motion with nothing in hand after a missed grasp; neither mode appears in any \name rollout. Fine-tuned $\pi_{0.5}$ fails predominantly by contact with the table and bins, and rarely progresses past a first item. \name also contacts the table and bins, but at backward offsets, over-shooting the items rather than stopping short of them. At the canonical pose on Book Stacking, \name grasps in 19/25 and stacks in 18/25 against 25/25 and 24--25/25 for Canonical-2D and fine-tuned $\pi_{0.5}$, a precision cost absent on Laundry Sort, where \name is best or tied-best at three of the four thresholds. Overall, we find that \name generalizes to perturbed base poses on real hardware more robustly than canonical training, explicit 3D representations, and large pretrained policies, with successful rollouts distributed across the sampled offset range rather than concentrated near the demonstration pose.

%% file: table/real.tex
\begin{table}[!t]
\centering
\small
\renewcommand{\arraystretch}{0.8}

\textbf{Laundry Sort}\\[2pt]
\begin{tabular*}{\linewidth}{l@{\extracolsep{\fill}}c@{\extracolsep{\fill}}c@{\extracolsep{\fill}}c@{\extracolsep{\fill}}c}
\toprule
Method & $\geq 1$ & $\geq 2$ & $\geq 3$ & $\geq 4$ \\
\midrule
\multicolumn{5}{l}{\textit{Randomized Pose}} \\
Canonical-2D            & 6/25  & 3/25  & 2/25  & 1/25 \\
DP3                     & 3/25  & 2/25  & 2/25  & \textbf{2/25} \\
Fine-tuned $\pi_{0.5}$  & 14/25 & 2/25  & 1/25  & 1/25 \\
\textbf{MobileVISTA (Ours)} & \textbf{18/25} & \textbf{12/25} & \textbf{6/25} & \textbf{2/25} \\
\midrule
\multicolumn{5}{l}{\textit{Canonical Pose}} \\
Canonical-2D            & 21/25 & \textbf{15/25} & 5/25  & 3/25 \\
DP3                     & 19/25 & 11/25 & 8/25  & 3/25 \\
Fine-tuned $\pi_{0.5}$  & 21/25 & 8/25  & 7/25  & \textbf{5/25} \\
\textbf{MobileVISTA (Ours)} & \textbf{24/25} & \textbf{15/25} & \textbf{9/25} & 4/25 \\
\bottomrule
\end{tabular*}

\vspace{6pt}
\textbf{Book Stacking}\\[2pt]
\begin{tabular*}{\linewidth}{l@{\extracolsep{\fill}}c@{\extracolsep{\fill}}c}
    \toprule
    \textbf{Method} & \textbf{Grasped} & \textbf{Stacked} \\
    \midrule
    \multicolumn{3}{l}{\textit{Randomized Pose}} \\
    Canonical-2D               & 1/25 & 1/25 \\
    DP3                          & 2/25 & 1/25 \\
    Fine-tuned $\pi_{0.5}$                   & 2/25 & 2/25 \\
    \textbf{MobileVISTA (Ours)}         & \textbf{17/25} & \textbf{13/25} \\
    \midrule
    \multicolumn{3}{l}{\textit{Canonical Pose}} \\
    Canonical-2D               & \textbf{25/25} & 24/25 \\
    DP3                          & 20/25 & 8/25 \\
    Fine-tuned $\pi_{0.5}$                    & \textbf{25/25} & \textbf{25/25} \\
    \textbf{\name (Ours)}         & 19/25 & 18/25 \\
    \bottomrule
\end{tabular*}

\caption{\small \textbf{Real-world success rates ($\uparrow$) for book stacking and laundry sort under canonical and randomized base poses.} For laundry sort, each cell reports the number of trials (of 25) in which at least $N$ items were correctly sorted. For book stacking, we report grasp and stack success in 25 trials. $r_{eval}=15$cm for randomized. \name maintains strong performance under base randomization, while baselines collapse despite strong canonical-pose performance.}
\label{tab:eval}
\end{table}

%% file: sec/5_conclusion.tex
\vspace{-2mm}
\section{Conclusion \& Limitations}
\vspace{-1mm}

\textbf{Limitations.} While MobileVISTA demonstrates strong viewpoint-robust policy learning, several limitations remain. First, synthesis quality depends on camera field-of-view: narrower FOVs reduce visible context, potentially degrading the consistency of generated views and policy performance. Second, we augment only egocentric cameras and not wrist-mounted cameras, leaving potential robustness gains from multi-view augmentation unexplored. Third, all results are bounded by the sampled perturbation range ($15$cm translation, $\pm5^\circ$ yaw), and behavior beyond it is untested.

\textbf{Conclusion.} We presented \name, a data generation framework that improves the robustness of mobile manipulation policies to base pose perturbations by jointly synthesizing pose-perturbed egocentric observations and retargeted actions from single-pose demonstrations. Unlike prior augmentation methods, which assume either a camera mounted off the actuated chain or non-trivial articulated robot geometry out of frame, our evaluation includes settings where neither assumption holds: the camera rides the actuated chain, and that chain occupies much of the observation. Experiments in both simulation and on a real Galaxea R1 Pro humanoid demonstrated \name's improved robustness over strong baselines. Promising future directions include integrating \name with large pretrained policies and extending augmentation to wrist-mounted cameras.

%% file: sec/appendix.tex
\clearpage
\appendix

\textbf{In addition to the following information, please see the supplementary video and project website (\url{https://sdwistreich.github.io/mobilevista/}) for a visual walkthrough of the framework and example rollout videos from each policy on each task.}

\section{Additional Details on the MobileVISTA Framework}

\subsection{Visual Observation Synthesis}
\label{sec:appendix_visual_synthesis}

\subsubsection{Robot Model and Segmentation}

We implement visual observation synthesis using MuJoCo via the robosuite framework. For each task, we render the robot using the same MuJoCo model used to instantiate the source demonstration, with non-robot geoms in the scene rendered invisible to produce a clean robot silhouette.

To extract a 2D mask of the robot, we use a different procedure across the simulated and real-world settings. In simulated experiments, we modify the scene's skybox color to a distinctive background color (red for the GR-1 humanoid setting, yellow for the bimanual Franka setting, the latter chosen to avoid conflicts with red environment objects), render the robot with all other geoms hidden, and threshold pixels by background color to produce a binary mask. In the real-world setting, we instead render the R1 Pro robot model in MuJoCo from a URDF specification and extract the robot silhouette directly from the resulting depth buffer (thresholded at $0.995$ to exclude background), then dilate the resulting mask by 5 pixels to capture edge artifacts.

We identify robot geoms in MuJoCo by body name prefixes: \texttt{robot0}/\texttt{gripper0} (and \texttt{robot1}/\texttt{gripper1} in the bimanual setting) for simulation, and the robot's named body geoms (e.g., \texttt{left\_arm\_link*}, \texttt{torso\_link*}) for the real-world R1 Pro setup. Rendering is performed at $512 \times 512$ resolution for simulated experiments and at the native dataset resolution for real-world experiments, upsampled by a factor of $6\times$ during point cloud reprojection.

\subsubsection{Depth Reprojection and Camera Extrinsics}

We lift the masked RGB-D observation into a colored point cloud using a pinhole camera model with intrinsics derived from the camera's vertical field of view ($\texttt{cam\_fovy}$ in MuJoCo). For real-world data, we use the ZED2 camera's calibrated intrinsics ($f_x = 58.4$, $f_y = 58.5$, $c_x = 74.7$, $c_y = 42.9$ at the dataset resolution, scaled with image dimensions during upsampling). Point clouds are constructed using Open3D's \texttt{RGBDImage.create\_from\_color\_and\_depth} with depth scale $1000$mm and a depth truncation of $5$m (simulation) or $10$m (real).

To lift the point cloud into the world frame, we use the camera-to-world extrinsics obtained from MuJoCo at the canonical configuration (\texttt{cam\_xpos}, \texttt{cam\_xmat}). After applying the base perturbation $\Delta p$, we re-render the robot model at its retargeted joint configuration and obtain new camera extrinsics from the resulting kinematic chain. We reproject the world-frame point cloud onto the new camera using an Open3D \texttt{OffscreenRenderer}, producing a pose-shifted RGB image and a corresponding hole mask from pixels not covered by reprojected points.

\subsubsection{Inpainting}

To fill missing regions in the reprojected RGB image, including both regions originally occupied by the segmented-out robot and newly disoccluded areas introduced by the base shift, we apply video inpainting with E2FGVI~\cite{liCvpr22vInpainting}. We process each trajectory's frames and per-frame hole masks as a temporal batch via E2FGVI's official pipeline. Prior to inpainting, we apply a $3 \times 3$ morphological closing (3 iterations) to the hole mask to fill small gaps introduced by point cloud sparsity. We use the model checkpoint \texttt{E2FGVI-HQ-CVPR22.pth}.

\subsubsection{Robot Re-Compositing}

After inpainting, we re-render the robot model at its retargeted joint configuration using MuJoCo, with non-robot geoms hidden as before. We composite the re-rendered robot onto the inpainted background by masking robot pixels from the rendered image. In the simulation pipeline, we use a color-based mask (excluding pixels matching the masking skybox color) to handle background bleed-through. In the real-world pipeline, we use the depth buffer from the re-rendered robot to identify robot pixels, again dilated by 5 pixels to capture edges. After compositing, real-world outputs are downsampled back to the original dataset resolution.

\subsection{Action Retargeting}
\label{sec:appendix_action_retargeting}

\subsubsection{Forward and Inverse Kinematics}

In the simulated humanoid (GR-1) and bimanual Franka settings, actions are originally represented as per-arm end-effector poses (3D position + 3D rotation vector per arm) in the robot base frame. Retargeting is performed in closed form by applying the inverse base perturbation transform to each arm's end-effector goal at each timestep (Section~\ref{sec:appendix_eef_perturbation}). In the bimanual setting, the rig caches both robot base poses and the shared camera pose; the yaw rotation is applied about the midpoint of the two base positions to preserve their relative configuration.

In the real-world (R1 Pro) setting, we instead solve for new joint configurations using whole-body inverse kinematics. We use the \texttt{mink}~\cite{zakka2024mink} IK solver with the following setup:
\begin{itemize}
    \item \textbf{FrameTasks} for the left and right end-effector sites (\texttt{left\_ee\_center}, \texttt{right\_ee\_center}), each with position cost and orientation cost set to $1.0$.
    \item \textbf{PostureTask} regularizing toward the previous configuration, with cost $10^{-5}$.
    \item \textbf{ConfigurationLimit} on joint positions.
   \item \textbf{VelocityLimit} with a per-step bound of $1.0$ on arm joints. The torso joint limit is set to $0.02$ for Book Stacking and $0.04$ for Laundry Sort, preferring arm motion over torso motion to maintain a stable viewpoint while allowing slightly more torso motion for the longer-horizon Laundry Sort task.
    \item \textbf{CollisionAvoidanceLimit} between end-effector and torso geom groups with a minimum collision distance of $0.04$m and a detection distance of $0.18$m.
\end{itemize}
We iterate the IK solver for up to 20 steps per frame with $\Delta t = 0.01$ using \texttt{quadprog} as the underlying solver, with damping $10^{-3}$ and a position tolerance of $0.01$m for convergence.

\subsubsection{End-Effector Goal Perturbation}
\label{sec:appendix_eef_perturbation}

Given a base perturbation $\Delta p = (\Delta b, \Delta \theta)$ with translation $\Delta b \in \mathbb{R}^2$ and yaw $\Delta \theta$, we construct the inverse transform $T_{\Delta p}^{-1}$ as a $4 \times 4$ matrix combining the inverse yaw rotation and a translation. For each arm's per-timestep end-effector position $\mathbf{p}_t$ (in the robot's base frame), the compensated position is $T_{\Delta p}^{-1} \cdot [\mathbf{p}_t; 1]$. Orientations, represented as rotation vectors, are updated by composing with the inverse yaw rotation: $R'_t = R_{\text{inv yaw}} \cdot R(\mathbf{r}_t)$, then converted back to a rotation vector. Gripper and finger commands are passed through unchanged.

For the bimanual Franka setting, the retargeting accounts for each arm's new base position after yaw is applied about the rig midpoint, so the per-arm compensation includes both a rotation and a translation derived from the change in that arm's base pose.

For the real-world R1 Pro setting, the end-effector position and orientation targets passed to IK at $t = 0$ are obtained by transforming the original end-effector poses around the rig pivot under $\Delta p$. For subsequent timesteps, IK is solved to track the original end-effector trajectory directly, allowing the arm to bend differently to reach the same world-frame target from the new base pose.

\subsection{Augmentation Sampling}
\label{sec:appendix_augmentation_sampling}

For each augmented trajectory, we sample a base translation offset $\Delta b$ uniformly within a disk of radius $r_{\text{train}} = 15$cm centered at the canonical pose using the standard formulation $(r, \theta) \sim (\sqrt{u} \cdot r_{\text{train}}, 2\pi v)$ for $u, v \sim \mathcal{U}(0, 1)$, with $\Delta b = (r \cos \theta, r \sin \theta, 0)$. No vertical offset is applied. When yaw jitter is enabled, we additionally sample $\Delta \theta \sim \mathcal{U}(-5^\circ, 5^\circ)$. We use the same training and evaluation radius $r_{\text{train}} = r_{\text{eval}} = 15$cm across all simulated and real-world tasks.

For simulated tasks, we generate one augmented trajectory per source demonstration, yielding 200 augmented trajectories from 200 source demonstrations. For real-world tasks, we generate ten augmented trajectories per source demonstration, yielding 500 augmented trajectories from 50 source demonstrations. We do not filter augmented trajectories based on IK convergence; in practice, the IK solver converges within tolerance for the vast majority of frames.

\section{Simulated Experiments}

\subsection{Task Definitions}
\label{sec:appendix_sim_tasks}

We evaluate on four tasks from DexMimicGen~\cite{jiang2024dexmimicen}. For the Can Sort task, we use an even split of expert demonstrations across can color (blue, red) for training. For the remainder of the tasks, we train using the first 200 demonstrations of the datasets and evaluate using the environment definitions provided by DexMimicGen, detailed below. Can Sort and Pouring are performed on a Fourier GR-1 humanoid, and Box and Drawer are performed on a bimanual Franka setup.

\begin{itemize}
    \item \textbf{Can Sort:} The robot must pick up a can with one arm, perform a handover to the other arm, and deposit it into the correct bin based on color. The handover phase is a coordination subtask requiring both arms to cooperate with synchronized trajectories. Can color and placement are randomized according to DexMimicGen defaults; the target bins are fixed.
    \item \textbf{Pouring:} A sequential task in which the robot must first pour a ball from a cup into a bowl using one arm, then pick up the bowl with the other arm and place it onto a green pad. The cup and bowl positions are randomized; the pad location is fixed.
    \item \textbf{Box:} The robot must pick up a box lid and place it onto the box using both arms simultaneously, requiring precise relative alignment between the two end-effectors. The positions of the box and lid are randomized.
    \item \textbf{Drawer:} A sequential task in which the robot must open a drawer with one arm before placing a mug inside with the other. The mug location is randomized according to DexMimicGen defaults.
\end{itemize}

\textbf{Camera setup.} For the bimanual Franka tasks (Box, Drawer), we use the \texttt{agentview} camera, a fixed rigid camera positioned between the two arms to approximate an egocentric viewpoint. For the GR-1 humanoid tasks (Can Sort, Pouring), we use the \texttt{robot0\_robotview} head-mounted camera. We widen its field of view from the DexMimicGen default of $50^\circ$ to $90^\circ$ to ensure the full manipulation workspace is visible.

\subsection{Tabled Per-Task Results}
\label{sec:appendix_detailed_results}

\textbf{Randomized Pose Perturbations.} \input{table/random_full} Table~\ref{tab:sim_detailed} reports full per-task success rates for all methods under the randomized base pose perturbation regime from Section~\ref{sec:sim} (Figure~\ref{fig:randomized}) of the main paper. Each cell reports mean success rate and standard deviation over three random seeds.

\textbf{Ablations.} \input{table/ablation_full} Table~\ref{tab:ablation_detailed} reports full per-task ablation results corresponding to Figure~\ref{fig:ablation} of the main paper. Each cell reports mean $\pm$ std over three random seeds. MobileVISTA wins across nearly all tasks and perturbation magnitudes. The exception is the Drawer task, where Fixed Robot Geom.\ is competitive or superior at intermediate perturbation radii, suggesting that this task may be less sensitive to inaccurate robot re-rendering.

\subsection{Comparison of MobileVISTA with Learned Novel View Synthesis}
\label{sec:appendix_psnr}

We provide additional details supporting the comparison between MobileVISTA and the learned novel view synthesis baseline GEN3C~\cite{ren2025gen3c} introduced in Section~\ref{sec:gen3c}. We report quantitative reconstruction quality, qualitative side-by-side comparisons, generation throughput, and downstream policy learning performance when training on views synthesized by each method.

\textbf{Data Generation.} We integrate GEN3C~\cite{ren2025gen3c} into our pipeline as a drop-in replacement for the geometry-based reprojection step in MobileVISTA, while keeping the action retargeting and robot re-compositing components unchanged. To ensure a fair comparison, we make several modifications to the default GEN3C setup.

First, rather than estimating depth and camera parameters from RGB using VIPE as in the official GEN3C implementation, we seed GEN3C with ground-truth depth and ground-truth camera intrinsics and extrinsics from the simulator. This removes a source of error that is orthogonal to the central question of how well each method synthesizes pose-shifted observations.

Second, we assume access to the same robot model used by MobileVISTA when rendering the perturbed scene from the new camera trajectory. This is necessary because, as discussed in Section~\ref{sec:retargeting}, a base perturbation induces a corresponding change in the egocentric camera trajectory through the base-to-camera kinematic chain after action retargeting, and the new camera trajectory cannot be computed without the robot model. Sharing this information with GEN3C ensures both methods use the same target camera trajectory for synthesis.

Finally, GEN3C generates video in fixed 121-frame chunks. We therefore generate each augmented trajectory in chunks of $121 \times 3 = 363$ frames; trajectories longer than this are truncated at the end. In practice, this truncation affected at most 12 frames per trajectory across the four tasks evaluated, and truncation always occurred at the trailing end of the trajectory.

\textbf{PSNR Comparison.} To quantify reconstruction quality, we render oracle egocentric views from the simulator at perturbed base poses and treat these as ground truth. We then synthesize views at the same perturbed poses using MobileVISTA and GEN3C and compute PSNR against the oracle. For each task, we sample 200 perturbed configurations within a disk of radius $r = 15$cm around the canonical pose. Table~\ref{tab:nvs_psnr} reports the mean PSNR across these configurations for each task.

\input{table/nvs_psnr}

\input{table/gen3c_time}

\begin{figure}[h]
    \centering
    \includegraphics[width=0.5\textwidth]{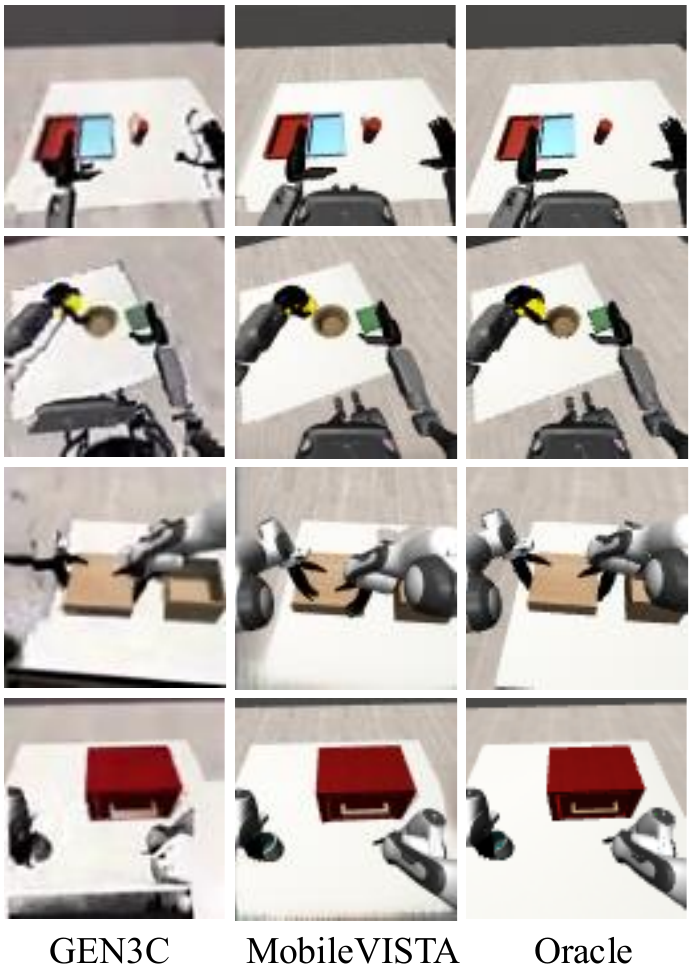}
    \caption{\small Qualitative comparison of synthesized egocentric observations. For each task, we show the simulator oracle view, the MobileVISTA-synthesized view, and the GEN3C-synthesized view at the same perturbed base pose. MobileVISTA more faithfully preserves scene geometry and object structure.}
    \label{fig:gen3c_qual}
\end{figure}

\vspace{5mm}

MobileVISTA produces higher PSNR than GEN3C across all four tasks, with the largest gaps appearing on the humanoid tasks (Can Sort and Pouring). The trends in Table~\ref{tab:nvs_psnr} are illustrated in the per-trajectory scatter visualization in Figure~\ref{fig:psnr_heatmap} of the main paper.

\textbf{Qualitative Comparisons.} Figure~\ref{fig:gen3c_qual} shows representative synthesized observations from MobileVISTA and GEN3C alongside the simulator oracle at matched perturbed base poses. MobileVISTA preserves sharp object boundaries and consistent scene geometry across perturbations, while GEN3C tends to blur object edges and distort surfaces under larger viewpoint shifts. These visual differences are consistent with the PSNR gap reported in Table~\ref{tab:nvs_psnr}.

\textbf{Data Generation Speed.} We measure the wall-clock time required to synthesize one augmented trajectory using each method, averaged across 200 trajectories per task. MobileVISTA timings are measured on a single NVIDIA L40S GPU, and GEN3C timings are measured on a single NVIDIA H200 GPU. Table~\ref{tab:timing} reports per-task averages. MobileVISTA generates augmented trajectories 4-7$\times$ faster than GEN3C across tasks. Note that as the H200 is a higher-end GPU than the L40S, so the reported speed advantage of MobileVISTA is a conservative estimate of the true gap on matched hardware. We postulate this gap arises as GEN3C performs iterative diffusion sampling for each frame, while MobileVISTA relies on a single forward pass through the geometry-based reprojection pipeline followed by inpainting.

\subsection{Policy Learning from GEN3C-Synthesized Views}
\label{sec:appendix_gen3c_policy}

We additionally evaluate downstream policy performance when training on GEN3C-synthesized views rather than MobileVISTA-synthesized views. To isolate the effect of the visual synthesis method, we pair GEN3C-synthesized observations with the same action retargeting procedure described in Section~\ref{sec:retargeting}, and train diffusion policies using the same configuration as MobileVISTA. We evaluate under the same randomized perturbation regime as Section~\ref{sec:sim}.

Results are shown in Figure~\ref{fig:gen3c_comparison}. The Canonical-2D, Simulator (Oracle), and MobileVISTA results are reproduced from Section~\ref{sec:sim} for reference. Policies trained on MobileVISTA-synthesized data outperform those trained on GEN3C-synthesized data across all four tasks and across the full range of evaluation perturbation magnitudes. We find the gap in downstream performance is consistent with the reconstruction quality gap reported in Table~\ref{tab:nvs_psnr}: tasks where the two methods produce more similar PSNR also generally show smaller gaps in policy success rate.

\section{Real-World Experiments}

\subsection{Hardware Setup}
\label{sec:appendix_real_hardware}

We deploy MobileVISTA and baselines on a Galaxea R1 Pro humanoid robot. The robot is equipped with a head-mounted ZED2 RGB-D camera providing the egocentric view used by all policies, and two wrist-mounted Intel RealSense RGB cameras.

To enable repeatable evaluation across base pose perturbations, we manually mark a grid on the floor extending $15$cm in each direction from the canonical robot base pose. Markings are placed by hand and verified against ground truth measurements before evaluation begins. During evaluation, we manually drive the robot to the sampled offset using these markings as reference, ensuring consistent and repeatable starting configurations across rollouts.

All policy inference is performed on a single NVIDIA RTX 4090 GPU.

\subsection{Data Collection \& Real World Task Definitions}
\label{sec:appendix_real_data}

\textbf{Teleoperation.} We collect expert demonstrations using the JoyLo teleoperation framework introduced in BEHAVIOR Robot Suite~\cite{jiang2025brs}. Demonstrations are recorded in ROS~2 at 10Hz with synchronized RGB-D images, proprioceptive states, and joint-space actions including arm, torso, and gripper commands.

\textbf{Book Stacking Task Definition.} The robot must grasp a vertically placed book from a designated location and reorient it horizontally onto an existing stack inside a bookshelf. We collect 50 expert demonstrations from the canonical (non-perturbed) robot pose, each approximately 150 timesteps in length.

\textbf{Laundry Sort Task Definition.} The robot must sort four laundry items into two bins based on category: two black socks must be placed into a grey bin on the left, and a yellow towel and a beige pair of shorts must be placed into a larger green bin on the right. We do not apply domain randomization to the item placement or appearance. This task is substantially more challenging than Book Stacking due to its longer horizon, the deformability of the items, and the bimanual coordination required to sort multiple objects in sequence.

\subsection{Evaluation Protocol}
\label{sec:appendix_real_eval}


For each trial, we sample a base translation offset uniformly within a disk of radius $r_{eval} = 15$cm centered at the canonical pose. We manually drive the robot to that offset using the floor markings described in Section~\ref{sec:appendix_real_hardware}. Yaw is not explicitly controlled, since policies are trained over the full $\pm 5^\circ$ yaw range used during augmentation.

From each sampled offset, we roll out all four policies (MobileVISTA, Canonical-2D, DP3, Fine-Tuned $\pi_{0.5}$) without repositioning the robot, enabling direct comparison from identical starting configurations. The order in which policies are evaluated at each offset is randomized and unknown to the experimenter to eliminate bias.

For Book Stacking, we score each rollout in two stages: Grasped, where the robot successfully grasps the book, and Stacked, where the book is successfully placed onto the existing stack. We perform 25 trials at the canonical pose and 25 trials at randomized poses, for a total of 50 trials per policy.

For Laundry Sort, we score each rollout by the number of items correctly sorted into the appropriate bin, ranging from $0$ to $4$. We perform 25 trials at the canonical pose and 25 trials at randomized poses, for a total of 50 trials per policy.



\section{Training and Inference Details}

\subsection{MobileVISTA and Canonical-2D Diffusion Policy}
\label{sec:appendix_dp}

We train MobileVISTA and Canonical-2D policies using Diffusion Policy~\cite{chi2023diffusionpolicy} within the Robomimic framework~\cite{robomimic2021}, using the official implementation\footnote{\url{https://github.com/real-stanford/diffusion_policy}}. All hyperparameters are identical across all tasks and all policy variants trained via this pipeline (MobileVISTA, Canonical-2D, Aug.\ Images Only, Aug.\ Actions Only, Fixed Robot Geom.). Full hyperparameters are listed in Table~\ref{tab:dp_hyperparams}. For simulated tasks we train for 1400 epochs. For the real-world Book Stacking task we train for 3500 epochs, and for the real-world Laundry task we train for 6500 epochs, where each epoch consists of 500 gradient steps.

The proprioceptive observations differ between simulation and real. In simulation, we use end-effector positions and quaternions for each arm, along with gripper joint positions. In the real-world setting, we instead use joint positions for the left and right arms, torso joint positions, and binary gripper open signals. All three RGB cameras (egocentric and two wrist cameras) are used in both settings.

\input{table/dp_hyperparams}

\subsection{DP3}
\label{sec:appendix_dp3}

We train DP3~\cite{Ze2024DP3} using the official implementation\footnote{\url{https://github.com/YanjieZe/3D-Diffusion-Policy}} with default hyperparameters. Key settings include: observation horizon 2, action horizon 8, prediction horizon 16, DDIM noise scheduler with 100 training steps and 10 inference steps, AdamW optimizer (lr 1e-4, weight decay 1e-6) with cosine scheduling and 500 warmup steps, batch size 128, and EMA (power 0.75). Point clouds are encoded with a PointNet encoder (output dim 64) using color as an additional input channel regardless of task. For simulated tasks we train for 1400 epochs. Due to compute constraints, real-world policies are trained by global step count rather than epochs: 634{,}845 steps for Book Stacking and 1.3M steps for Laundry. We train three random seeds per task for all simulated experiments.

\subsection{Fine-Tuned $\pi_{0.5}$}
\label{sec:appendix_pi05}

We fine-tune $\pi_{0.5}$~\cite{intelligence2025pi05visionlanguageactionmodelopenworld} using the official \texttt{openpi} repository\footnote{\url{https://github.com/Physical-Intelligence/openpi}}, initializing from the released \texttt{pi05\_base} checkpoint. We use LoRA fine-tuning (gemma\_2b\_lora for the vision-language backbone and gemma\_300m\_lora for the action expert), with EMA disabled. All other training settings follow the \texttt{pi05\_libero} reference config in the \texttt{openpi} repository, with \texttt{action\_dim=32} and \texttt{action\_horizon=16}.

\textbf{Simulated experiments.} For each task, we train three random seeds for up to 30{,}000 steps, saving a checkpoint every 1{,}000 steps. For each seed, we sweep over all saved checkpoints by evaluating 15 rollouts at each of three perturbation radii ($r \in \{0, 5, 10\}$cm) and selecting the checkpoint with the highest mean success rate across all three radii. We then run full evaluation (50 rollouts per radius, all perturbation magnitudes) on the selected checkpoint for each seed, and report results averaged over the three seeds.

\textbf{Real-world experiments.} For the real-world Book Stacking task, we average the selected checkpoint steps across the three best-performing simulated seeds to obtain a single fine-tuning step count of 22{,}000 steps. For the real-world Laundry task, we scale this by the ratio of Robomimic diffusion policy training durations between the two real-world tasks ($6500 / 3500 \approx 1.86\times$), yielding 41{,}000 steps. This scaling aligns the relative fine-tuning budget of $\pi_{0.5}$ with that of the Robomimic-based diffusion policies across the two tasks.

%% file: table/random_full.tex
\begin{table*}[!t]
\centering
\footnotesize
\setlength{\tabcolsep}{5pt}
\renewcommand{\arraystretch}{1}
\small
\begin{tabular}{llccccccc}
\toprule
Task & Method & $r{=}0$ & $r{=}1$ & $r{=}3$ & $r{=}5$ & $r{=}7$ & $r{=}10$ & $r{=}15$ \\
\midrule
\multirow{5}{*}{Can Sort}
& Canonical-2D            & $0.480_{\pm 0.02}$ & $0.500_{\pm 0.09}$ & $0.393_{\pm 0.04}$ & $0.440_{\pm 0.05}$ & $0.327_{\pm 0.03}$ & $0.240_{\pm 0.00}$ & $0.120_{\pm 0.06}$ \\
& DP3                     & $0.733_{\pm 0.08}$ & $0.713_{\pm 0.06}$ & $0.600_{\pm 0.13}$ & $0.407_{\pm 0.05}$ & $0.327_{\pm 0.06}$ & $0.153_{\pm 0.04}$ & $0.060_{\pm 0.02}$ \\
& Fine-tuned $\pi_{0.5}$  & $0.926_{\pm 0.05}$ & $0.906_{\pm 0.01}$ & $0.833_{\pm 0.06}$ & $0.827_{\pm 0.10}$ & $0.780_{\pm 0.09}$ & $0.607_{\pm 0.09}$ & $0.320_{\pm 0.03}$ \\
& \textbf{MobileVISTA (Ours)} & $\mathbf{0.953_{\pm 0.03}}$ & $\mathbf{0.967_{\pm 0.02}}$ & $\mathbf{0.933_{\pm 0.02}}$ & $\mathbf{0.887_{\pm 0.05}}$ & $\mathbf{0.873_{\pm 0.05}}$ & $\mathbf{0.793_{\pm 0.06}}$ & $\mathbf{0.527_{\pm 0.04}}$ \\
\cmidrule(lr){2-9}
& \textit{Simulator (Oracle)} & $\textit{0.913}_{\pm 0.02}$ & $\textit{0.927}_{\pm 0.05}$ & $\textit{0.893}_{\pm 0.01}$ & $\textit{0.927}_{\pm 0.03}$ & $\textit{0.873}_{\pm 0.05}$ & $\textit{0.727}_{\pm 0.01}$ & $\textit{0.527}_{\pm 0.08}$ \\
\midrule
\multirow{5}{*}{Pouring}
& Canonical-2D            & $0.547_{\pm 0.03}$ & $0.413_{\pm 0.01}$ & $0.460_{\pm 0.08}$ & $0.353_{\pm 0.03}$ & $0.200_{\pm 0.09}$ & $0.080_{\pm 0.02}$ & $0.073_{\pm 0.06}$ \\
& DP3                     & $0.287_{\pm 0.08}$ & $0.240_{\pm 0.09}$ & $0.247_{\pm 0.06}$ & $0.173_{\pm 0.06}$ & $0.113_{\pm 0.03}$ & $0.080_{\pm 0.06}$ & $0.013_{\pm 0.01}$ \\
& Fine-tuned $\pi_{0.5}$  & $0.467_{\pm 0.03}$ & $0.447_{\pm 0.02}$ & $0.433_{\pm 0.08}$ & $0.333_{\pm 0.05}$ & $0.207_{\pm 0.06}$ & $0.093_{\pm 0.03}$ & $0.013_{\pm 0.01}$ \\
& \textbf{MobileVISTA (Ours)} & $\mathbf{0.673_{\pm 0.02}}$ & $\mathbf{0.640_{\pm 0.03}}$ & $\mathbf{0.640_{\pm 0.06}}$ & $\mathbf{0.613_{\pm 0.04}}$ & $\mathbf{0.607_{\pm 0.10}}$ & $\mathbf{0.547_{\pm 0.12}}$ & $\mathbf{0.460_{\pm 0.02}}$ \\
\cmidrule(lr){2-9}
& \textit{Simulator (Oracle)} & $\textit{0.773}_{\pm 0.03}$ & $\textit{0.787}_{\pm 0.06}$ & $\textit{0.713}_{\pm 0.13}$ & $\textit{0.727}_{\pm 0.03}$ & $\textit{0.673}_{\pm 0.03}$ & $\textit{0.700}_{\pm 0.07}$ & $\textit{0.560}_{\pm 0.11}$ \\
\midrule
\multirow{5}{*}{Box}
& Canonical-2D            & $0.307_{\pm 0.11}$ & $0.273_{\pm 0.02}$ & $0.227_{\pm 0.09}$ & $0.147_{\pm 0.10}$ & $0.100_{\pm 0.03}$ & $0.047_{\pm 0.02}$ & $0.000_{\pm 0.00}$ \\
& DP3                     & $0.407_{\pm 0.01}$ & $0.473_{\pm 0.02}$ & $0.373_{\pm 0.05}$ & $0.333_{\pm 0.02}$ & $0.293_{\pm 0.05}$ & $0.213_{\pm 0.03}$ & $0.087_{\pm 0.05}$ \\
& Fine-tuned $\pi_{0.5}$  & $0.187_{\pm 0.10}$ & $0.187_{\pm 0.06}$ & $0.187_{\pm 0.08}$ & $0.147_{\pm 0.01}$ & $0.107_{\pm 0.01}$ & $0.060_{\pm 0.03}$ & $0.020_{\pm 0.00}$ \\
& \textbf{MobileVISTA (Ours)} & $\mathbf{0.520_{\pm 0.14}}$ & $\mathbf{0.600_{\pm 0.07}}$ & $\mathbf{0.427_{\pm 0.08}}$ & $\mathbf{0.440_{\pm 0.07}}$ & $\mathbf{0.413_{\pm 0.03}}$ & $\mathbf{0.393_{\pm 0.07}}$ & $\mathbf{0.247_{\pm 0.05}}$ \\
\cmidrule(lr){2-9}
& \textit{Simulator (Oracle)} & $\textit{0.313}_{\pm 0.09}$ & $\textit{0.373}_{\pm 0.02}$ & $\textit{0.433}_{\pm 0.10}$ & $\textit{0.407}_{\pm 0.06}$ & $\textit{0.473}_{\pm 0.05}$ & $\textit{0.373}_{\pm 0.06}$ & $\textit{0.333}_{\pm 0.01}$ \\
\midrule
\multirow{5}{*}{Drawer}
& Canonical-2D            & $0.287_{\pm 0.10}$ & $0.260_{\pm 0.05}$ & $0.280_{\pm 0.00}$ & $0.100_{\pm 0.04}$ & $0.120_{\pm 0.03}$ & $0.040_{\pm 0.02}$ & $0.013_{\pm 0.01}$ \\
& DP3                     & $0.260_{\pm 0.00}$ & $0.273_{\pm 0.02}$ & $0.220_{\pm 0.05}$ & $0.180_{\pm 0.03}$ & $0.093_{\pm 0.05}$ & $0.067_{\pm 0.06}$ & $0.040_{\pm 0.02}$ \\
& Fine-tuned $\pi_{0.5}$  & $0.580_{\pm 0.09}$ & $0.513_{\pm 0.05}$ & $0.400_{\pm 0.02}$ & $0.373_{\pm 0.11}$ & $0.227_{\pm 0.06}$ & $0.167_{\pm 0.08}$ & $0.060_{\pm 0.02}$ \\
& \textbf{MobileVISTA (Ours)} & $\mathbf{0.653_{\pm 0.04}}$ & $\mathbf{0.620_{\pm 0.05}}$ & $\mathbf{0.580_{\pm 0.07}}$ & $\mathbf{0.593_{\pm 0.06}}$ & $\mathbf{0.587_{\pm 0.02}}$ & $\mathbf{0.487_{\pm 0.09}}$ & $\mathbf{0.413_{\pm 0.01}}$ \\
\cmidrule(lr){2-9}
& \textit{Simulator (Oracle)} & $\textit{0.720}_{\pm 0.03}$ & $\textit{0.780}_{\pm 0.07}$ & $\textit{0.627}_{\pm 0.03}$ & $\textit{0.713}_{\pm 0.07}$ & $\textit{0.567}_{\pm 0.05}$ & $\textit{0.640}_{\pm 0.09}$ & $\textit{0.393}_{\pm 0.03}$ \\
\bottomrule
\end{tabular}
\caption{Detailed per-task success rates under randomized pose perturbation across evaluation radii $r_{eval} \in \{0, 1, 3, 5, 7, 10, 15\}$cm with yaw $\Delta\theta \sim \mathcal{U}(-5^\circ, 5^\circ)$. Each cell reports the mean $\pm$ std over three random seeds (50 evaluation rollouts per seed). \textbf{Bold} indicates the best non-oracle method per row. Simulator (Oracle) is reported as a reference upper bound.}
\label{tab:sim_detailed}
\end{table*}

%% file: table/ablation_full.tex
\begin{table*}[!t]
\centering
\renewcommand{\arraystretch}{1}
\setlength{\tabcolsep}{3pt}
\resizebox{\linewidth}{!}{%
\begin{tabular}{llccccccc}
\toprule
Task & Method & $r{=}0$ & $r{=}1$ & $r{=}3$ & $r{=}5$ & $r{=}7$ & $r{=}10$ & $r{=}15$ \\
\midrule
\multirow{5}{*}{Can Sort}
& Canonical-2D          & $0.48 \pm 0.02$ & $0.50 \pm 0.09$ & $0.39 \pm 0.04$ & $0.44 \pm 0.05$ & $0.33 \pm 0.03$ & $0.24 \pm 0.00$ & $0.12 \pm 0.06$ \\
& Aug.\ Images Only     & $0.54 \pm 0.05$ & $0.45 \pm 0.09$ & $0.36 \pm 0.00$ & $0.17 \pm 0.10$ & $0.11 \pm 0.04$ & $0.03 \pm 0.02$ & $0.03 \pm 0.02$ \\
& Aug.\ Actions Only    & $0.08 \pm 0.07$ & $0.11 \pm 0.08$ & $0.09 \pm 0.01$ & $0.07 \pm 0.06$ & $0.07 \pm 0.05$ & $0.07 \pm 0.04$ & $0.07 \pm 0.05$ \\
& Fixed Robot Geom.     & $0.80 \pm 0.05$ & $0.80 \pm 0.07$ & $0.81 \pm 0.05$ & $0.64 \pm 0.05$ & $0.46 \pm 0.09$ & $0.31 \pm 0.10$ & $0.17 \pm 0.06$ \\
& \textbf{MobileVISTA (Ours)} & $\mathbf{0.95 \pm 0.03}$ & $\mathbf{0.97 \pm 0.02}$ & $\mathbf{0.93 \pm 0.02}$ & $\mathbf{0.89 \pm 0.05}$ & $\mathbf{0.87 \pm 0.05}$ & $\mathbf{0.79 \pm 0.06}$ & $\mathbf{0.53 \pm 0.04}$ \\
\midrule
\multirow{5}{*}{Pouring}
& Canonical-2D          & $0.55 \pm 0.03$ & $0.41 \pm 0.01$ & $0.46 \pm 0.08$ & $0.35 \pm 0.03$ & $0.20 \pm 0.09$ & $0.08 \pm 0.02$ & $0.07 \pm 0.06$ \\
& Aug.\ Images Only     & $0.40 \pm 0.05$ & $0.41 \pm 0.04$ & $0.33 \pm 0.12$ & $0.31 \pm 0.08$ & $0.08 \pm 0.02$ & $0.04 \pm 0.04$ & $0.05 \pm 0.02$ \\
& Aug.\ Actions Only    & $0.00 \pm 0.00$ & $0.01 \pm 0.01$ & $0.00 \pm 0.00$ & $0.01 \pm 0.01$ & $0.01 \pm 0.01$ & $0.00 \pm 0.00$ & $0.00 \pm 0.00$ \\
& Fixed Robot Geom.     & $0.31 \pm 0.08$ & $0.25 \pm 0.01$ & $0.16 \pm 0.04$ & $0.10 \pm 0.07$ & $0.04 \pm 0.02$ & $0.03 \pm 0.01$ & $0.01 \pm 0.02$ \\
& \textbf{MobileVISTA (Ours)} & $\mathbf{0.67 \pm 0.02}$ & $\mathbf{0.64 \pm 0.03}$ & $\mathbf{0.64 \pm 0.06}$ & $\mathbf{0.61 \pm 0.04}$ & $\mathbf{0.61 \pm 0.10}$ & $\mathbf{0.55 \pm 0.12}$ & $\mathbf{0.46 \pm 0.02}$ \\
\midrule
\multirow{5}{*}{Box}
& Canonical-2D          & $0.31 \pm 0.11$ & $0.27 \pm 0.02$ & $0.23 \pm 0.09$ & $0.15 \pm 0.10$ & $0.10 \pm 0.03$ & $0.05 \pm 0.02$ & $0.00 \pm 0.00$ \\
& Aug.\ Images Only     & $0.15 \pm 0.07$ & $0.13 \pm 0.08$ & $0.05 \pm 0.01$ & $0.02 \pm 0.00$ & $0.01 \pm 0.01$ & $0.01 \pm 0.01$ & $0.00 \pm 0.00$ \\
& Aug.\ Actions Only    & $0.01 \pm 0.01$ & $0.01 \pm 0.01$ & $0.00 \pm 0.00$ & $0.00 \pm 0.00$ & $0.00 \pm 0.00$ & $0.01 \pm 0.01$ & $0.00 \pm 0.00$ \\
& Fixed Robot Geom.     & $0.22 \pm 0.07$ & $0.25 \pm 0.09$ & $0.23 \pm 0.02$ & $0.22 \pm 0.11$ & $0.22 \pm 0.05$ & $0.11 \pm 0.03$ & $0.11 \pm 0.03$ \\
& \textbf{MobileVISTA (Ours)} & $\mathbf{0.52 \pm 0.14}$ & $\mathbf{0.60 \pm 0.07}$ & $\mathbf{0.43 \pm 0.08}$ & $\mathbf{0.44 \pm 0.07}$ & $\mathbf{0.41 \pm 0.03}$ & $\mathbf{0.39 \pm 0.07}$ & $\mathbf{0.25 \pm 0.05}$ \\
\midrule
\multirow{5}{*}{Drawer}
& Canonical-2D          & $0.29 \pm 0.10$ & $0.26 \pm 0.05$ & $0.28 \pm 0.00$ & $0.10 \pm 0.04$ & $0.12 \pm 0.03$ & $0.04 \pm 0.02$ & $0.01 \pm 0.01$ \\
& Aug.\ Images Only     & $0.12 \pm 0.03$ & $0.09 \pm 0.01$ & $0.07 \pm 0.05$ & $0.01 \pm 0.01$ & $0.01 \pm 0.01$ & $0.01 \pm 0.01$ & $0.00 \pm 0.00$ \\
& Aug.\ Actions Only    & $0.01 \pm 0.01$ & $0.01 \pm 0.02$ & $0.01 \pm 0.01$ & $0.00 \pm 0.00$ & $0.00 \pm 0.00$ & $0.01 \pm 0.01$ & $0.00 \pm 0.00$ \\
& Fixed Robot Geom.     & $0.55 \pm 0.12$ & $\mathbf{0.62 \pm 0.09}$ & $\mathbf{0.64 \pm 0.07}$ & $\mathbf{0.59 \pm 0.13}$ & $\mathbf{0.66 \pm 0.05}$ & $\mathbf{0.51 \pm 0.08}$ & $0.36 \pm 0.05$ \\
& \textbf{MobileVISTA (Ours)} & $\mathbf{0.65 \pm 0.04}$ & $\mathbf{0.62 \pm 0.05}$ & $0.58 \pm 0.07$ & $\mathbf{0.59 \pm 0.06}$ & $0.59 \pm 0.02$ & $0.49 \pm 0.09$ & $\mathbf{0.41 \pm 0.01}$ \\
\bottomrule
\end{tabular}%
}
\caption{Detailed ablation results across tasks. Success rate (mean $\pm$ std over three seeds) vs. evaluation radius $r_{eval} \in \{0, 1, 3, 5, 7, 10, 15\}$cm with yaw $\Delta\theta \sim \mathcal{U}(-5^\circ, 5^\circ)$. \textbf{Bold} indicates the best method per column within each task.}
\label{tab:ablation_detailed}
\end{table*}

%% file: table/nvs_psnr.tex
\begin{table*}[!t]
\centering
\small
\setlength{\tabcolsep}{6pt}
\begin{tabular}{lccc|ccc|ccc|ccc}
    \toprule
    & \multicolumn{3}{c}{\textbf{Can Sort}} 
    & \multicolumn{3}{c}{\textbf{Pouring}} 
    & \multicolumn{3}{c}{\textbf{Box}} 
    & \multicolumn{3}{c}{\textbf{Drawer}} \\
    \cmidrule(lr){2-4} \cmidrule(lr){5-7} \cmidrule(lr){8-10} \cmidrule(lr){11-13}
    \textbf{Method} 
    & 0-5cm & 5-10cm & 10-15cm
    & 0-5cm & 5-10cm & 10-15cm
    & 0-5cm & 5-10cm & 10-15cm
    & 0-5cm & 5-10cm & 10-15cm \\
    \midrule
    GEN3C & 16.29 & 14.42 & 13.08
           & 17.74 & 15.09 & 13.28
           & 16.37 & 13.75 & 11.78
           & 16.32 & 15.76 & 14.35 \\
    \textbf{M.Vista} & \textbf{25.02} & \textbf{24.05} & \textbf{23.28}
                             & \textbf{23.71} & \textbf{23.25} & \textbf{22.18}
                             & \textbf{16.47} & \textbf{16.01} & \textbf{14.72}
                             & \textbf{18.67} & \textbf{18.93} & \textbf{18.29} \\
    \bottomrule
\end{tabular}
\caption{$\uparrow$ PSNR (dB) between synthesized egocentric views and simulator oracle images across base perturbation magnitudes $r$.}
\label{tab:nvs_psnr}
\end{table*}

%% file: table/gen3c_time.tex
\setlength{\columnsep}{8pt}
\begin{wraptable}[9]{r}{0.22\textwidth}
    \centering
    \small
    \setlength{\tabcolsep}{3pt}
    \renewcommand{\arraystretch}{0.9}
    \begin{tabular}{lcc}
        \toprule
        & \textbf{GEN3C} & \textbf{M.Vista} \\
        \midrule
        Can Sort & 14.93 & \textbf{3.81} \\
        Pouring  & 13.65 & \textbf{4.31} \\
        Box      & 14.93 & \textbf{2.39} \\
        Drawer   & 16.85 & \textbf{2.99} \\
        \bottomrule
    \end{tabular}
    \captionsetup{width=0.95\linewidth}
    \caption{\footnotesize \textbf{$\downarrow$ Avg. synthesis time per trajectory (minutes).} \name is 4-7$\times$ faster than GEN3C across tasks.}
    \label{tab:timing}
\end{wraptable}

%% file: table/dp_hyperparams.tex
\begin{table}[t]
\centering
\normalsize
\begin{tabular}{ll}
\toprule
Hyperparameter & Value \\
\midrule
\multicolumn{2}{l}{\textit{Optimization}} \\
Learning rate & 1e-4 \\
L2 regularization & 0.0 \\
Batch size & 16 \\
Steps per epoch & 500 \\
Action normalization & min-max \\
\midrule
\multicolumn{2}{l}{\textit{Policy horizons}} \\
Observation horizon & 2 \\
Action horizon & 8 \\
Prediction horizon & 16 \\
\midrule
\multicolumn{2}{l}{\textit{Diffusion}} \\
Noise scheduler & DDIM \\
Training diffusion steps & 100 \\
Inference diffusion steps & 10 \\
Beta schedule & \texttt{squaredcos\_cap\_v2} \\
Prediction type & $\epsilon$-prediction \\
EMA power & 0.75 \\
\midrule
\multicolumn{2}{l}{\textit{UNet}} \\
Channel dims & [256, 512, 1024] \\
Kernel size & 5 \\
Groups & 8 \\
Diffusion step embed dim & 256 \\
\midrule
\multicolumn{2}{l}{\textit{Image encoder}} \\
Backbone & ResNet-18 (no pretraining) \\
Pooling & SpatialSoftmax (32 keypoints) \\
Feature dim per camera & 64 \\
Input resolution & $84 \times 84$ \\
Random crop & $76 \times 76$ \\
\bottomrule
\end{tabular}
\caption{Diffusion Policy hyperparameters used for all MobileVISTA and Canonical-2D policy variants across simulated and real-world tasks.}
\label{tab:dp_hyperparams}
\end{table}